\documentclass[10pt,twocolumn,letterpaper]{article}
\usepackage{authblk}
\usepackage[pagenumbers]{cvpr} 

\definecolor{cmblu}{RGB}{51,102,240}
\definecolor{cmred}{RGB}{241,22,22}
\definecolor{cgreen}{RGB}{68,152,80}
\definecolor{corange}{RGB}{236,99,63}

\usepackage[dvipsnames]{xcolor}

\usepackage{graphicx}

\usepackage{epsfig}
\usepackage{bbding} 
\usepackage{pifont} 
\usepackage{wrapfig} 
\usepackage{microtype}
\usepackage{amsfonts}

\usepackage{amsmath,amssymb} 
\usepackage{mathrsfs} 
\usepackage{algorithmic,algorithm}
\usepackage{url} 
\usepackage{multirow}  
\usepackage{pifont} 
\usepackage{microtype} 
\usepackage{array}
\usepackage{soul}
\usepackage{booktabs}
\usepackage{caption,subcaption}
\usepackage{colortbl}

\newtheorem{theorem}{Theorem}

\newcommand{\cmark}{\textcolor{cgreen}{\ding{51}}}%
\newcommand{\xmark}{\textcolor{corange}{\ding{55}}}%

\newcommand{\modelshortname}{GUES\xspace}

\newcommand{\setting}{OMG-DA\xspace}

\definecolor{cvprblue}{rgb}{0.21,0.49,0.74}
\usepackage[pagebackref,breaklinks,colorlinks,allcolors=cvprblue]{hyperref}
\def\paperID{16156} 
\def\confName{CVPR}
\def\confYear{2025}

\title{Domain Adaptive Diabetic Retinopathy Grading with Model \\Absence and Flowing Data}

\author[1]{Wenxin Su}
\author[1,2]{Song Tang\thanks{Corresponding author}}
\author[3]{Xiaofeng Liu}
\author[4]{Xiaojing Yi}
\author[6]{Mao Ye}
\author[1]{\\Chunxiao Zu}
\author[5]{Jiahao Li}
\author[7]{Xiatian Zhu}
\affil[1]{University of Shanghai for Science and Technology \textsuperscript{2} Universität Hamburg}
\affil[3]{Yale University \textsuperscript{4} Sichuan Eye Hospital \textsuperscript{5} Peking Union Medical College Hospital}
\affil[6]{University of Electronic Science and Technology of China \textsuperscript{7} University of Surrey}

\affil[ ]{
{\tt\small steventangsong@gmail.com}
}

\begin{document}
\maketitle
\begin{abstract}
Domain shift (the difference between source and target domains) poses a significant challenge in clinical applications, e.g., Diabetic Retinopathy (DR) grading. Despite considering certain clinical requirements, like source data privacy, conventional transfer methods are predominantly model-centered and often struggle to prevent model-targeted attacks. In this paper, we address a challenging {\bf O}nline {\bf M}odel-a{\bf G}nostic {\bf D}omain {\bf A}daptation ({\bf \setting}) setting, driven by the demands of clinical environments. 
This setting is characterized by the absence of the model and the flow of target data.   
To tackle the new challenge, we propose a novel approach, {\bf G}enerative {\bf U}nadversarial {\bf E}xample{\bf S} ({\bf GUES}), which enables adaptation from a data-centric perspective.  
Specifically, we first theoretically reformulate conventional perturbation optimization in a generative way---learning a perturbation generation function with a latent input variable. 
During model instantiation, we leverage a Variational AutoEncoder to express this function. 
The encoder with the reparameterization trick predicts the latent input, whilst the decoder is responsible for the generation.
Furthermore, the saliency map is selected as pseudo-perturbation labels. 
Because it not only captures potential lesions but also theoretically provides an upper bound on the function input, enabling the identification of the latent variable. 
Extensive comparative experiments on DR benchmarks with both frozen pre-trained models and trainable models demonstrate the superiority of {\modelshortname}, showing robustness even with small batch size.

\end{abstract}    
\section{Introduction}
\label{sec:intro}
Diabetic Retinopathy (DR) is a significant health concern, ranking among the leading causes of blindness and affecting millions of people worldwide~\cite{abdelmaksoud2020comprehensive}. Early-stage intervention for DR is crucial to preserve vision, highlighting the importance of timely diagnosis~\cite{singer1992screening}. Although deep learning (DL) has demonstrated promising results in automating the grading of DR~\cite{wu2020coarse,he2020cabnet,dai2021deep}, deploying DL models in real-world clinical settings remains challenging. For example, DL models often struggle to generalize effectively to complex scenarios, such as variations in imaging equipment, ethnic groups, or temporal factors, leading to different data distributions, a challenge known as domain shift~\cite{kouw2018introduction}. This issue significantly hampers the widespread adoption and success of DL-based diagnostic tools in clinical practice~\cite{li2021applications}.
\begin{figure}[t] 
    \setlength{\belowcaptionskip}{0pt}
    \setlength{\abovecaptionskip}{0pt}
    \begin{center}
     \includegraphics[width=0.95\linewidth]{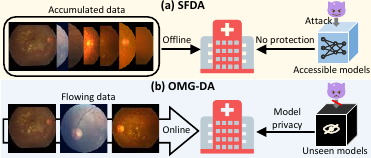}
    \end{center}
    \caption{
    Comparison between the OMG-DA and SFDA settings. (a) In SFDA, adaptation builds upon the accumulated data, which demands significant storage resources in the hospital. Additionally, the models' architecture and parameters are accessible, exposing them to potential attacks. (b) {\setting} provides a practical scenario: Flowing data mimic the patients' arrival in a stream way, and the models are unseen (strictly controlled) before using them, avoiding attacks like membership inference attacks~\cite{shokri2017membership}. 
    }
    \label{fig:setting}
\end{figure}
\begin{table*}[t]
        \caption{Comparison between different transfer settings. Notation: source (s), target (t), data $\boldsymbol{x}$, label $\boldsymbol{y}$, loss function ${L}\left(\cdot\right)$.} 
	\label{tab:setting}
	\renewcommand\tabcolsep{2.5pt}
	\renewcommand\arraystretch{1.0}
	\scriptsize
	\centering
	\begin{tabular}{ l | c |c |c || c | c | c | c}
	\toprule
	{Setting name}  &Model availability &Data flow &Source data privacy &Source data &Target data &Train loss &Test loss  \\
        \midrule
        Fine-tuning (FT)                       &\xmark &\xmark &\cmark &-- &$\boldsymbol{x}_t,\boldsymbol{y}_t$&${L}\left(\boldsymbol{x}_t,\boldsymbol{y}_t\right)$ &-- \\
        Domain generalization (DG)             &\xmark &\cmark &\xmark &$\boldsymbol{x}_s,\boldsymbol{y}_s$ &- &${L}\left(\boldsymbol{x}_s,\boldsymbol{y}_s\right)$ &-\\
        Unsupervised domain adaptation (UDA)   &\xmark &\xmark &\xmark &$\boldsymbol{x}_s$,$\boldsymbol{y}_s$ &$\boldsymbol{x}_t$ &${L}\left(\boldsymbol{x}_s,\boldsymbol{y}_s\right)+{L}\left(\boldsymbol{x}_t\right)$ &--\\
        Source-free domain adaptation (SFDA)   &\xmark &\xmark &\cmark&-- &$\boldsymbol{x}_t$ &${L}\left(\boldsymbol{x}_t\right)$ &--  \\
        Test-time adaptation (TTA)            &\xmark &\cmark &\cmark&-- &$\boldsymbol{x}_t$ &-- &${L}\left(\boldsymbol{x}_t\right)$  \\
        Online model-agnostic domain adaptation ({\setting})  &\cmark &\cmark &\cmark &-- &$\boldsymbol{x}_t$ &-- &-- \\
		\bottomrule
\end{tabular}
\label{tab:setting}
\end{table*}

Recently, many adaptation methods for grading diabetic retinopathy (DR) have focused on addressing the issue of domain shift~\cite{zhang2022diabetic,nguyen2021self,ran2024source,che2023towards}. The initial focus on classic transfer learning strategies, including Unsupervised Domain Adaptation (UDA)~\cite{nguyen2021self} and Domain Generalization (DG)~\cite{che2023towards,atwany2022drgen}, necessitated the availability of well-annotated source data. Nevertheless, the growing emphasis on privacy protection has shifted research toward the Source-Free Domain Adaptation (SFDA) framework~\cite{ran2024source,zhang2022diabetic}. SFDA involves adapting a source model—pre-trained on the source domain—to the target domain in a self-supervised manner, thereby ensuring the protection of source patient data. 

In recent developments, specific needs have emerged in the clinical field. 
The introduction of model weight-based techniques for reconstructing training data has created a demand for {\em model privacy}~\cite{yin2021see,shokri2017membership}, which goes beyond traditional source data protection. In addition, there is a growing requirement for models capable of handling incoming patient data in a flowing fashion, referred to as a {\em flowing data} constraint~\cite{valanarasu2024fly,wang2020tent}. 
Unfortunately, existing SFDA methods cannot effectively address this challenge, as they rely on full access to the model and require offline training on a pre-collected dataset. 
Fig.~\ref{fig:setting} provides an intuitive illustration of this issue.

In this paper, we consider a clinically motivated setting, called {\em {\bf O}nline {\bf M}odel-a{\bf G}nostic {\bf D}omain {\bf A}daptation} ({\bf \setting}) and propose a novel {\em {\bf G}enerative {\bf U}nadversarial {\bf E}xample{\bf S}} ({\bf \modelshortname}) approach for the DR grading problem in this new setting. 
Specifically, {\setting} presents an extreme safety scenario: The available target data is unlabeled and arrives in a flowing format, with no prior information about the pre-trained source model and data. 
Tab.~\ref{tab:setting} provides a detailed comparison with previous adaptation settings.


In {\modelshortname}, we address the absence of source data and pre-trained models by producing generalized unadversarial examples~\cite{salman2021unadversarial} for unlabeled target data. 
To this end, we introduce generative unadversarial learning, which theoretically reformulates conventional iterative perturbation optimization. 
This new method aims to learn a generative function for perturbations and involves addressing two key tasks: (1) Identifying the latent function input, which is the derivative of initial random noise w.r.t. image data, and (2) selecting a self-supervised property to serve as pseudo-perturbation labels. 
In practice, we leverage a Variational Autoencoder (VAE) ~\cite{kingma2013auto} based approach to facilitate this learning process. 
In terms of function representation, we model the latent input using the encoder along with the reparameterization trick, whilst the decoder accomplishes the generation of individual-perturbation. 
Additionally, we choose the saliency map as the pseudo-perturbation label for two reasons: (1) It helps discover potential lesions, and (2) it aids in identifying the latent input by providing an upper bound.

Our {\bf contributions} are summarized as follows:
\begin{itemize}
\item Pioneering a novel transfer setting {\setting}, which is closer to real-world clinical scenarios and meets three typical requirements at the same time, including (1) model absence, (2) flowing data, and (3) source data privacy. 



\item Developing a new {\setting} approach {\modelshortname} in the context of DR grading, grounded on the generative unadversarial examples theory, where we learn an individual-perturbation generative function under saliency map supervision, removing relying on labels and models.


\item Extensive evaluations on four DR benchmarks, indicating that {\modelshortname} can largely promote the source model's performance in the target domain, as well as trainable test-time adaptation models, even at small batch size.

\end{itemize}

\section{Related work}\label{sec:rewk}

\paragraph{Adaptation methods for DR grading.}
Driven by real-world medical requirements, domain adaptation has been an attractive topic in this DR grading issue. 
For instance, Nguyen et al.~\cite{nguyen2021self} introduce a UDA approach that enables the model to focus on vessel structures that remain invariant to domain shifts via image reconstruction using labeled source domain data. In SFDA, Zhang et al.~\cite{zhang2022diabetic} propose generating labeled, target-style retinal images to improve the source model's generalization, relying solely on a pre-trained source model and unlabeled target images. Additionally, DG in DR grading has been explored through domain-invariant feature learning approaches~\cite{atwany2022drgen} and divergence-based methods~\cite{che2023towards}, leveraging labeled source domain data. 

These methods above rely on labeled data, require full access to the model, and necessitate offline training on a pre-collected dataset. In real clinical settings, these requirements can be impractical due to constraints around data and model privacy, as well as the need for real-time adaptability without retraining. In contrast, GUES provides an online adaptation solution that operates without requiring labeled data or access to the model, addressing the domain adaptation problem in DR grading.

\vspace{0.1cm}
\noindent{\bf Unadversarial learning.} 
Unadversarial learning was initially developed by Salman et al.~\cite{salman2021unadversarial}, aiming to modify input image distribution to make them more easily recognizable by the model. 
Current mainstreams achieve this learning process by adding class-specific perturbations to the input images. 
Here, the perturbations are generated based on the gradient of an objective function w.r.t. image. 
This approach allows for the design of unadversarial examples without model training.
For example, based on~\cite{salman2021unadversarial}, NSA~\cite{sharma2023nsa} introduces a method to generate more natural perturbations using a trainable generator. Similarly, CAT~\cite{liu2023cat} demonstrates a new distance metric for generating unadversarial examples.

All existing unadversarial learning methods require access to model parameters, outputs, and labeled data. This dependency invalidates them in our \setting setting which only has access to unlabeled target data. 
In addition to this, our {\modelshortname} produces individualized unadversarial examples in a generative manner, which stands out from previous methods that focus on class-specific unadversarial examples.

\vspace{0.1cm}
\noindent{\bf Saliency map for medical image.} 
A fine-grained saliency map is a pixel feature generated by calculating the central-surround differences within images, identifying salient regions without any need for training~\cite{montabone2010human}. 
This feature is widely applied in various medical image analysis tasks to extract pathologically important regions~\cite{tomar2024visual,wei2024saliency,qiu2024augpaste,huang2024ssit}. 
For instance, in DR grading, studies such as~\cite{huang2024ssit, qiu2024augpaste} utilize saliency maps to guide models in focusing on critical features like the optic disc, cup, and vessel structures. Similarly, in brain tumor detection, Tomar et al.~\cite{tomar2024visual} leverage saliency maps to enhance the model’s attention on tumor and bone structures. In skin cancer detection, saliency maps help isolate lesion regions with distinctive features, such as lumpiness, which are essential for accurate diagnosis~\cite{wei2024saliency}.

As stated above, existing works primarily use saliency maps to highlight lesion regions. 
Unlike the conventional usage of saliency maps, {\modelshortname} selects saliency maps as pseudo-perturbation labels.


\section{Problem Statement of {\setting}}
Given two different but related domains, \textit{i.e.}, source domain $\mathcal{S}$ and target domain $\mathcal{T}$, $\mathcal{S}$ contains $n_s$ labeled samples, while $\mathcal{T}$ has $n$ unlabeled data. 
Both labeled and unlabeled samples share the same $C$ categories. 
Let $\mathcal{X}_s$ and $\mathcal{Y}_s$ be the source samples and the corresponding labels. Similarly, we denote the target samples and their labels by $\mathcal{X}_t=\{{\boldsymbol{x}^{i}_t\}_{i=1}^{n}}$ and $\mathcal{Y}_t=\{{y}^{i}_{t}\}_{i=1}^{n}$, respectively, where $n$ signifies the number of samples. 
The source model $\theta_{s}$ is  pre-trained on $\{\mathcal{X}_s, \mathcal{Y}_s\}$. 

{\setting} is featured in (1) the absence of the source model $\theta_{s}$ and domain $\mathcal{S}$ and (2) the flowing target data $\mathcal{X}_t$ the same as the TTA setting~\cite{wang2020tent}. 
Unlike previous transfer settings that are model adaptation-centered~\cite{tang2024source,tang2024unified,liang2020we,yang2021nrc}, {\setting} considers adaptation from the perspective of data. Specifically, {\setting} aims to modify the distribution of target data to facilitate downstream tasks.

\section{Methodology}\label{sec:method}

\begin{figure*}[t]
    \begin{center}
        \includegraphics[width=0.95\linewidth]{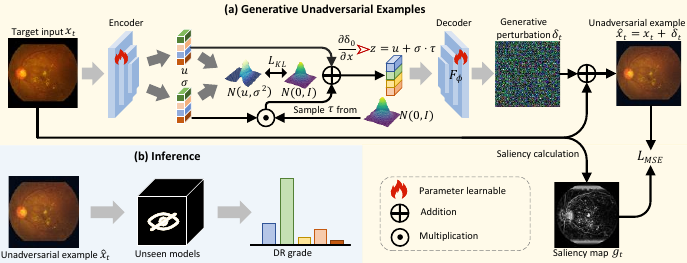}
    \end{center}
    \caption{
    The instantiation framework of {\modelshortname} in the {\setting} setting. 
    (a) For target input $x_t$, the VAE model generates individual perturbation $\delta_t={F}_{\Phi}\left(\frac{\partial \delta_0}{\partial x} \right)$. After that, the by-pass path incorporates $\delta_t$ and $x_t$ to create the generative unadversarial example $\hat{x}_t$. Treating $x_t$'s saliency map $g_t$ as reconstruction supervision for model training.   
    (b) At the inference phase, the generated unadversarial example $\hat{x}_t$ is directly provided to the frozen source model or other trainable models. 
    }
    \label{fig:fw}
\end{figure*}
\begin{figure}[t]
\setlength{\belowcaptionskip}{0pt}
\setlength{\abovecaptionskip}{0pt}
\begin{center}
    \includegraphics[width=0.9\linewidth]{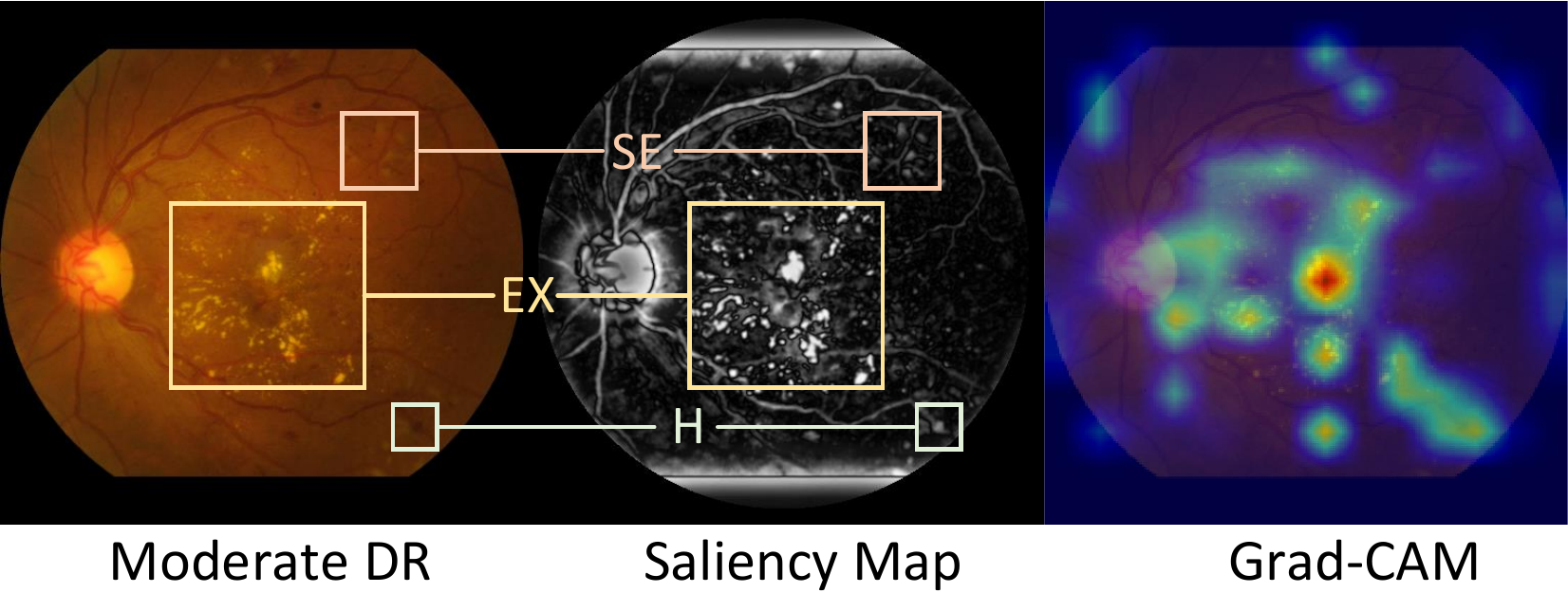}
\end{center}
\caption{Explanation in choosing fine-grained saliency maps as supervision. 
{\bf Left:} The testing fundus image selected from ``Moderate DR" class in APTOS demonstrates that H (hemorrhages), SE (soft exudates), and EX (hard exudates) are essential characteristics to judge the DR grade. {\bf Middle:} The saliency map highlights those lesions.
{\bf Right:} The gradient-CAM visualization of the source model on task DDR$\to$APTOS.}  
\label{fig:sal_sample}
\end{figure}

\subsection{Generative Unadversarial Examples}
{The part begins with a brief recap of traditional unadversarial learning~\cite{salman2021unadversarial}.}
Unlike adversarial learning~\cite{szegedy2013intriguing} that generates confusing samples to mislead models, unadversarial learning aims to construct generalized samples, tackling promoting out-of-distribution issues. Formally, this learning can be summarized in the following optimization problem. 
\begin{equation}
    \label{eqn:unadv}
    \begin{aligned}
        \hat{\delta} = \arg\min \limits_{\delta} {L}({f}_{\theta}({x+\delta}), y),
        s.t.~||\delta|| \leq \epsilon\\
    \end{aligned}
\end{equation} 
where ${L}\left(\cdot\right)$ denotes objective function, e.g., cross-entropy loss for classification tasks, $x$ and $y$ are input image and its label, ${f}_{\theta}$ is a pre-trained model with parameters $\theta$, $\delta$ is a perturbation, $\epsilon$ is a small threshold. 
The current scheme solves this problem in an iterative way formulated as 
\begin{equation}
    \label{eqn:solve-intera}
    \small
    \begin{aligned}
        \delta_{k+1} = \delta_{k} + \alpha\cdot{\rm{sign}}\left(\nabla_{x}{L}({f}_{\theta}({x+\delta_{k}}), y)\right), 
        k \in [0,K-1],
    \end{aligned} 
\end{equation}
where $\alpha$ is a trade-off parameter, $K$ is iteration number, $\delta_0$ is an initial random noise. 
In the inference phase, the optimal perturbation $\hat{\delta}$ is integrated into the input $x$, forming an unadversarial example $\hat{x}= x+\hat{\delta}$, which is easily recognizable by the model ${f}_{\theta}$.
Obviously, the conventional unadversarial paradigm cannot meet our {\setting} setting due to the absence of $f_{\theta}$, $L$, and the label $y$.

In this paper, we re-consider the iterative optimization process above and obtain the theorem below (The proof is provided in \texttt{Supplementary}). 
\begin{theorem}
\textit{Given the unadversarial learning problem defined in Eq.~\eqref{eqn:unadv}, the iterative process featured by Eq.~\eqref{eqn:solve-intera} can be expressed as the following generative form.} 
\begin{equation}
    \label{eqn:theo1}
    {\delta}_{k} = {\delta_0} + V \cdot {F}_{\Phi}\left(\frac{\partial \delta_0}{\partial x} \right), 
\end{equation}
where $\delta_0$ is an initial random noise, $V>0$ is a bound constant, $F_{\Phi}$ is a generative function, $\frac{\partial \delta_0}{\partial x}$ is a latent variable. 
\label{thm-one} 
\end{theorem}
\vspace{-0.02cm}
Grounded on Theorem~\ref{thm-one}, we have: When ${\delta}_{k}$ converges to optimal $\hat{\delta}$, i.e., ${\delta}_{k}\to\hat{\delta}$, function $F_{\Phi}$ also evolves to the optimal one, denoted by $\hat{F}_{\Phi}$, i.e., $F_{\Phi}\to\hat{F}_{\Phi}$. 
This provides an insight: 
{\em The unadversarial learning problem above can also be solved in a generative fashion.} 
Correspondingly, the generated data are termed generative unadversarial examples.

\subsection{Model Instantiation} \label{sec:nhd}
Within this context of generative unadversarial learning, conventional unadversarial learning presented in Eq.~\eqref{eqn:unadv} is boiled down to learning $F_{\Phi}\left(\frac{\partial \delta_0}{\partial x} \right)$. 
We can achieve this by training a generative neural network.     
To make this solution sense, we have to solve two difficulties as follows. 
{\bf (A)} One is the identification of $\frac{\partial \delta_0}{\partial x}$ when the relationship between them is unknown. 
{\bf (B)} The other is selecting property supervision (pseudo-perturbation labels) to drive ${\delta}_{k}\to\hat{\delta}$. 

\vspace{0.1cm}
\noindent{\bf Solution to problem A.} In practice, considering derivative $\frac{\partial \delta_0}{\partial x}$ is relevant with both random noise $\delta_0$ and input image $x$, we sample it from a certain Gaussian distribution associated with ${x}$. 
Furthermore, the output size of $F_{\Phi}$ is the same as ${x}$. 
Therefore, we employ the VAE model to jointly model $\frac{\partial \delta_0}{\partial x}$ and $F_{\Phi}$, since VAE is an autoencoder characterized by random sampling.  
Specifically, as shown in Fig.\ref{fig:fw} (a), we approximate $\frac{\partial \delta_0}{\partial x}$ by latent variable $z$, which is jointly determined by input $x$ and sampled random signal $\tau$.    
As for $F_{\Phi}$, it is demonstrated by the decoder module. 
Suppose $D(\cdot)$ is the decoder, $E_{\tau}(\cdot)$ is the encoder with reparameterization trick, our scheme can be formulated as 
\begin{equation}
    \label{eqn:pro-a}
    \small
    \begin{aligned}
    {D}(E_{\tau}(x))\to{F}_{\Phi}\left(\frac{\partial \delta_0}{\partial x} \right),~  
    E_{\tau}(x)\to\frac{\partial \delta_0}{\partial x},~D(\cdot)\to{F}_{\Phi}(\cdot).
    \end{aligned}
\end{equation}

\noindent{\bf Solution to problem B.} We adopt the fine-grained saliency map as the supervision. 
Two reasons contribute to our selection. 
First of all, the empirical results show that, for the specific task of DR grading, the saliency map is an acceptable pseudo-perturbation. 
Specifically, perturbation in the unadversarial context enhances the regions associated with the category and reduces the prominence of other areas, thereby identifying the lesion zones relevant to DR grading.   
As illustrated in Fig.~\ref{fig:sal_sample}, the saliency map effectively identifies potential lesions, such as hemorrhages, soft exudates, and hard exudates. Furthermore, it includes gradient information, as it highlights regions similar to Grad-CAM (Right side).
More importantly, we have the theorem below. (The proof is provided in \texttt{Supplementary})   
\begin{theorem}
\textit{Given the partial derivatives of the initial random noise $\delta_0$ w.r.t image $x$ is $\frac{\partial \delta_0}{\partial x}$ and $x$'s saliency map is $s=G(x)$ where $G(\cdot)$ is the computation function of the saliency map. We have the following relationship:   
} 
\begin{equation}
    \label{eqn:theo2}
    \frac{\partial \delta_0}{\partial x} \leq U \cdot s, 
\end{equation}
where $U>0$ is a bound constant. 
\label{thm-two} 
\end{theorem}
Theorem~\ref{thm-two} suggests that the saliency map provides upper bounds for $\frac{\partial \delta_0}{\partial x}$. 
Namely, $s$ provides relaxed descriptions for the variation $\frac{\partial \delta_0}{\partial x}$.
This can help guide the learning of $\frac{\partial \delta_0}{\partial x}$.



\vspace{0.1cm}     
\noindent{\bf {\modelshortname} framework.}
Based on the analysis above, we instantiate {\modelshortname} as the framework depicted in Fig.~\ref{fig:fw}.
As shown in sub-figure (a), our method integrates a VAE model and by-pass connection, achieving the learning of ${F}_{\Phi}\left(\frac{\partial \delta_0}{\partial x} \right)$. Specifically, $\frac{\partial \delta_0}{\partial x}$ is sampled from an input $x_t$-featured Gaussian distribution $N(\mu, \sigma)$, which is jointly learned using the encoder and reparameterization. The decoder, representing ${F}_{\Phi}$, then transforms $\frac{\partial \delta_0}{\partial x}$ into a generative perturbation $\delta_t$. Finally, the by-pass structure incorporates $x_t$ and $\delta_t$ to produce $\hat{x}_t$.
During the inference phase, as shown in Fig.~\ref{fig:fw} (b), for a specific testing sample, the trained {\modelshortname} model outputs the corresponding generative unadversarial examples to the frozen or fine-tuning model that early unseen.

\vspace{0.1cm}
\noindent{\bf Loss function}
The loss function for {\modelshortname} training consists of two components. 
First, we enforce the latent space with mean $\mu$ and variance $\sigma$ satisfy the standard normal distribution $\mathcal{N}(0,I)$. Suppose the encoder in VAE models the posterior distribution $q(z|x_t)=\mathcal{N}(\mu(x_t),\sigma^2(x_t))$, this regularization can be formulated as:  
\begin{equation}
    \begin{aligned}
    L_\mathrm{KL} &= D_{\mathrm{KL}} \left(q(z|x_t) \| \mathcal{N}(0,I)\right),
    \end{aligned}
    \label{eqn:kl}
\end{equation}
where function $D_{\mathrm{KL}}$ computes the Kullback-Leibler divergence. 
The other reconstruction loss between the unadversarial example $\hat{x}_t$ and saliency map $g_t$ is presented by the following regression form: 
\begin{equation}
    \begin{aligned}
        L_{\text{MSE}} &= \| \hat{x}_t - {g}_t \|_2.
    \end{aligned}
    \label{eqn:mse}
\end{equation}
 Formally, combining Eq.~\eqref{eqn:kl} and Eq.~\eqref{eqn:mse}, the final objective of {\modelshortname} can be summarized as:
\begin{equation} 
    {L_{{\text{GUES}}}} = \alpha{L_{{\text{KL}}}} + \beta{L_{{\text{MSE}}}},
    \label{eqn:loss-all}
\end{equation}
where $\alpha$ and $\beta$ are trade-off parameters. For clarity, we summarize the training procedure of {\modelshortname} in Algorithm~\ref{alg:algorithm}.

In Eq.~\eqref{eqn:loss-all}, the first item $L_{\rm{KL}}$ ensures the learning of $\frac{\partial \delta_0}{\partial x}$. 
On the one hand, as aforementioned, we use $z$ to present $\frac{\partial \delta_0}{\partial x}$. 
$L_{\rm{KL}}$ aligns the $z$ space with $\mathcal{N}(0,I)$, thereby linking the random noise to $\frac{\partial \delta_0}{\partial x}$. On the other hand, $q(z|x_t)$ in $L_{\rm{KL}}$ is a function of input $x_t$, building relationship $x_t$ to  $\frac{\partial \delta_0}{\partial x}$. 
Additionally, the reconstruction regulated by the seconded item $L_{\rm{MSE}}$ encourage ${\delta}_{k}\to\hat{\delta}$.

\begin{algorithm}[t]
    \caption{The pipeline of proposed {\modelshortname}}
    \label{alg:algorithm}
    \raggedright
    \textbf{Input}: Online batch samples $\mathcal{B}$, a trainable VAE $\theta_v$ consists of an encoder $E_\tau(\cdot)$ with the reparameterization trick and a decoder $D(\cdot)$.\\
    
    \textbf{Procedure}:
    \begin{algorithmic}[1] 
    \FOR {$x_i$ in $\mathcal{B}$}
    \STATE Approximate the latent function input by $E_{\tau}(x_i)$;

    \STATE Learn a generative function for perturbation $\delta_i$ by $D$;
    \STATE Calculate individual perturbations $\delta_i$ by ${D}(E_{\tau}(x))$;
     \STATE Create the unadversarial example $\hat{x}_i$ by incorporating $\delta_i$ and $x_i$ through the bypass path.
    \STATE Generate a fine-grained saliency map $g_i$ of the $x_i$;
    \STATE Update $\theta_v$ with Eq.\eqref{eqn:loss-all}, taking $g_t$ as a supervision.
    \ENDFOR
    \STATE \textbf{return} The generative unadversarial examples $\hat{x}$  
    \end{algorithmic}
\end{algorithm}

\section{Experiments}
\label{sec:experiments}

\subsection{Datasets}
We perform evaluation experiments on four existing fundus benchmarks, including 
\textbf{APTOS}~\cite{APTOS2019}, 
\textbf{DDR}~\cite{li2019diagnostic},
\textbf{DeepDR}~\cite{liu2022deepdrid},
and \textbf{Messidor-2} (termed MD2)~\cite{decenciere2014feedback}. 
Those datasets share five grading/classes: no DR, mild DR, moderate DR, severe DR, and proliferative DR. Taking each dataset as a separate domain, we form 12 transfer tasks crossing domains. 
For example, as APTOS is the source domain while the others are target domains, we have three transfer tasks APTOS$\to$DDR, APTOS$\to$DeepDR, and APTOS$\to$MD2. 
The illustration of the label distribution and the domain shift of the four datasets is demonstrated in \texttt{Supplementary}. 

It should be {\bf noted that} all datasets have a severe class imbalance (e.g., ``no DR" class itself takes up to 45.8\% of the DDR dataset).

\begin{table*}[t]
	\caption{ The results of Source, SFDA, TTA, OMG-DA, and OMG-DA combination methods on datasets APTOS, DeepDR, DDR, and MD2 are presented. The improvements over baseline methods Source, SHOT-IM, and TENT are highlighted as ({\color{cmred}+x.x}).}
	\label{tab:oh-st}
	\renewcommand\tabcolsep{1.2pt}
	\renewcommand\arraystretch{1.05}
	\scriptsize
	\centering
	\begin{tabular}{ l l | c c c| c c c| c c c|c c c| c c c| c c c|c c c}
        \toprule
        \multirow{2}{*}{Method} &\multirow{2}{*}{Venue} 
        &\multicolumn{3}{c|}{APTOS$\to$DDR} &\multicolumn{3}{c|}{APTOS$\to$DeepDR} &\multicolumn{3}{c|}{APTOS$\to$MD2} &\multicolumn{3}{c|}{DDR$\to$APTOS}
        &\multicolumn{3}{c|}{DDR$\to$DeepDR} &\multicolumn{3}{c|}{DDR$\to$MD2}&\multicolumn{3}{c}{DeepDR$\to$APTOS}  \\
         &&ACC &QWK &AVG &ACC &QWK &AVG &ACC &QWK &AVG &ACC &QWK &AVG &ACC &QWK &AVG &ACC &QWK &AVG &ACC &QWK &AVG\\
        \midrule
        Source  &-- & 60.6 &59.2&59.9 &52.6	&71.7&62.1 &60.9&\textbf{\color{cmblu}48.7}&54.8 &65.6 &72.9 &69.3 &45.0	&60.6 &52.8&49.4 &34.6 &42.0 &43.4	&71.6	&57.5 \\
        \rowcolor{gray! 20}{\bf {\modelshortname}}   &--  \cellcolor{gray!20} &62.0	\cellcolor{gray!20}&59.5	&60.8
        &53.0	&69.7	&61.3
        &59.8	&46.7	&53.3
        &76.0	&81.8	&78.9
        &56.4	&68.7	&62.5
        &59.1	&47.6	&53.3
        &46.5	&74.1	&60.3
        \\

	\midrule 
        SHOT~\cite{liang2020we}   &ICML20  &66.9 	&69.0 	&67.9 
        &\textbf{\color{cmblu}53.6} 	&\textbf{\color{cmblu}73.5} 	&\textbf{\color{cmblu}63.6}
        &51.7 	&38.0 	&44.8 
        &77.0 	&\textbf{\color{cmblu}84.2} 	&\textbf{\color{cmblu}80.6}
        &59.2 	&74.6 	&66.9 
        &57.1 	&43.1 	&50.1 &\textbf{\color{cmblu}62.3} 	&82.5 	&72.4
         \\
        NRC~\cite{yang2021nrc}    &NeurIPS21  & 61.9	&65.2	&63.5
        &51.6	&70.9	&61.3
        &54.1	&40.3	&47.2
        &60.3	&76.3	&68.3
        &52.0	&69.1	&60.5
        &50.6	&35.2	&42.9 &52.0	&74.3	&63.2\\
        CoWA~\cite{lee2022confidence}   &ICML22  &59.0	&64.9	&62.0
        &51.0	&70.7	&60.8 
        &53.0	&37.3	&45.1
        &57.2	&74.5	&65.8
        &50.1	&66.8	&58.4
        &53.1	&39.6	&46.3 &50.4	&73.0	&61.7\\
        PLUE~\cite{Litrico_2023_CVPR}   &CVPR23  &62.0 	&65.1 	&63.6 
        &51.3 	&69.7 	&60.5 
        &54.5 	&41.1 	&47.8 
        &63.4 	&64.2 	&63.8 
        &54.3 	&64.8 	&59.5 
        &51.6 	&28.2 	&39.9 &54.6 	&70.5 	&62.6\\
        TPDS~\cite{tang2024source}   &IJCV24  &66.6 	&67.8 	&67.2 
        &51.6 	&71.5 	&61.5 
        &52.7 	&40.7 	&46.7 
        &76.6 	&83.9 	&80.2 
        &58.0 	&73.1 	&65.6 
        &54.5 	&41.0 	&47.8 &60.8 	&80.0 	&70.4
        \\
        \midrule
        SHOT-IM~\cite{liang2020we} &ICML20 &66.5 	&\textbf{\color{cmblu}69.2} 	&67.9 
        &52.6 	&73.6 	&63.1 
        &53.2 	&36.6 	&44.9 
        &75.9 	&82.1 	&79.0 
        &58.5 	&73.9 	&66.2 
        &57.3 	&43.5 	&50.4 
        &61.9 	&\textbf{\color{cmblu}84.0} 	&\textbf{\color{cmblu}72.9}
        \\
        TENT~\cite{wang2020tent} &ICLR20 &59.9 	&50.2 	&55.1 
        &53.1 	&70.1 	&61.6 
        &\textbf{\color{cmblu}61.4} &48.4 &\textbf{\color{cmblu}54.9}  
        &75.2 	&82.4 	&78.8 
        &55.1 	&68.4 	&61.7 
        &60.8 	&50.5 	&55.6 
        &60.2 	&79.7 	&69.9

        \\
        SAR ~\cite{niutowards} &ICLR23&67.9 	&63.8 	&65.8 
        &\textbf{\color{cmblu}53.6} 	&73.0 	&63.3 
        &57.2 	&44.9 	&51.1 
        &75.6 	&83.5 	&79.5 
        &55.6 	&71.6 	&63.6 
        &49.2 	&37.0 	&43.1 
        &59.5 	&79.3 	&69.4 
        \\
        \midrule

        \rowcolor{gray! 20}{\bf {\modelshortname+SHOT-IM}}   &--   &\textbf{\color{cmblu}68.6} 	&68.5 	&\textbf{\color{cmblu}68.5}
        &53.5 	&72.8 	&63.2 
        &55.7 	&43.8 	&49.7 
        &\textbf{\color{cmblu}77.2} 	&83.1 	&80.2 
        &\textbf{\color{cmblu}60.5} 	&\textbf{\color{cmblu}75.1} 	&\textbf{\color{cmblu}67.8}
        &61.5 	&51.2 	&56.3 &62.6 &83.2 	&\textbf{\color{cmblu}72.9}

        \\
        \rowcolor{gray! 20}{\bf {\modelshortname+TENT}}   &--   &61.8 	&56.3 	&59.0 
        &53.2 	&70.0 	&61.6 
        &61.1 	&47.2 	&54.1 
        &75.9 	&83.0 	&79.4 
        &58.7 	&70.8 	&64.7 
        &\textbf{\color{cmblu}63.3} 	&\textbf{\color{cmblu}53.8} 	&\textbf{\color{cmblu}58.6} 
        &54.9 	&77.6 	&66.3 
        
         \\
        
	\end{tabular}\\

	\begin{tabular}{ l l | c c c | c c c| c c c| c c c| c c c| c c c}
        \midrule
        \multirow{2}{*}{Method} &\multirow{2}{*}{Venue} 
        &\multicolumn{3}{c|}{DeepDR$\to$DDR}
        &\multicolumn{3}{c|}{DeepDR$\to$MD2} &\multicolumn{3}{c|}{MD2$\to$APTOS} &\multicolumn{3}{c|}{MD2$\to$DDR} &\multicolumn{3}{c|}{MD2$\to$DeepDR}
        &\multicolumn{3}{c}{Avg.} \\
	& &ACC &QWK &AVG &ACC &QWK &AVG &ACC &QWK &AVG &ACC &QWK &AVG &ACC &QWK &AVG &ACC &QWK &AVG\\
        \midrule
	Source &-- 
        &56.4	&66.9	&61.7
        &48.7	&50.2	&49.4
        &43.9	&70.3	&57.1
        &60.2	&56.5	&58.3
        &59.8	&58.4	&59.1
        &53.9	&60.1	&57.0

        \\	
    
        \rowcolor{gray! 20}{\bf {\modelshortname}}  &--  
        &57.3	&65.6	&61.4
        &48.3	&52.0	&50.1
        &59.3	&76.2	&67.7
        &64.7	&55.8	&60.2
        &58.6	&57.1	&57.8
        &58.4 ({\color{cmred}+4.5})	&62.9 ({\color{cmred}+2.8})	&60.7 ({\color{cmred}+3.7})\\
        \midrule
	SHOT~\cite{liang2020we}    &ICML20 
        &57.4 	&71.2 	&64.3 
        &48.8 	&41.5 	&45.2 
        &52.7 	&73.0 	&62.8 
        &54.6 	&59.2 	&56.9 
        &59.6 	&\textbf{\color{cmblu}70.2} 	&\textbf{\color{cmblu}64.9}
        &58.4 	&65.0 	&61.7 
        \\
        NRC~\cite{yang2021nrc}     &NeurIPS21 
        &44.9	&60.9	&52.9
        &49.8	&41.8	&45.8
        &48.8	&69.1	&58.9
        &52.8	&52.8	&52.8
        &58.0	&62.8	&60.4 
        &53.1	&59.9	&56.5\\
        CoWA~\cite{lee2022confidence}    &ICML22  
        &48.7	&58.4	&53.6
        &49.9	&42.7	&46.3
        &51.0	&70.1	&60.5
        &49.6	&50.9	&50.3
        &57.6	&60.6	&59.1 
        &56.9	&61.9	&59.4\\
        PLUE~\cite{Litrico_2023_CVPR}    &CVPR23  
        &47.2 	&53.5 	&50.4 
        &56.4 	&47.4 	&51.9
        &56.0 	&69.1 	&62.6 
        &56.5 	&54.3 	&55.4 
        &58.8  &64.8 	&61.8 &55.5 &57.7 	&56.6 \\
        TPDS~\cite{tang2024source}    &IJCV24  
            &59.3 	&69.4 	&64.3 
            &50.5 	&42.4 	&46.4 
            &60.3 	&74.9 	&67.6 
            &60.0 	&60.4 	&60.2 
            &58.9 	&63.0 	&60.9 
            &59.2 	&64.0 	&61.6 
            \\
        \midrule
        SHOT-IM~\cite{liang2020we} &ICML20 
            &54.6 	&69.4 	&62.0 
        &51.2 	&38.2 	&44.7 
        &61.6 	&77.9 	&69.7 
        &57.0 	&58.7 	&57.9 
        &57.5 	&69.8 	&63.7 
        &59.0 	&64.7 	&61.9 
         \\
        
        TENT~\cite{wang2020tent} &ICLR20 
        &58.5 	&45.4 	&51.9 
        &58.3 	&56.5 	&57.4
        &55.1 	&74.1 	&64.6 
        &55.8 	&31.7 	&43.7 
        &58.0 	&53.6 	&55.8 
        &59.3 	&59.2 	&59.3

         \\

        SAR~\cite{niutowards} &ICLR23 
        &53.0 	&66.3 	&59.6 
        &42.6 	&33.1 	&37.9 
        &55.2 	&73.0 	&64.1 
        &49.7 	&48.3 	&49.0 
        &56.7 	&65.8 	&61.3 
        &56.3 	&61.6 	&59.0 
        \\
        \midrule

        \rowcolor{gray! 20} {\bf {GUES+SHOT-IM}}  &--   
        &62.3 	&\textbf{\color{cmblu}71.5} 	&\textbf{\color{cmblu}66.9}
        &52.8 	&47.8 	&50.3 
        &\textbf{\color{cmblu}62.8} 	&\textbf{\color{cmblu}78.1} 	&\textbf{\color{cmblu}70.4}
        &\textbf{\color{cmblu}66.0} 	&\textbf{\color{cmblu}59.4} 	&\textbf{\color{cmblu}62.7}
        &\textbf{\color{cmblu}60.6}  	&68.6  	&64.6  
        &\textbf{\color{cmblu}62.0} ({\color{cmred}+3.0}) 	&\textbf{\color{cmblu}66.9} ({\color{cmred}+2.2})
        &\textbf{\color{cmblu}64.5} ({\color{cmred}+2.6})

        \\
        \rowcolor{gray! 20} {\bf {GUES+TENT}}  &-- 
        &\textbf{\color{cmblu}62.6} 	&61.4 	&62.0 
        &\textbf{\color{cmblu}59.1}	&\textbf{\color{cmblu}57.4}	&\textbf{\color{cmblu}58.2}
        &59.3	&75.6	&67.5
        &64.0	&50.5	&57.2
        &58.3	&56.0	&57.1
        &61.0 ({\color{cmred}+1.7})	&63.3 ({\color{cmred}+4.1})	&62.2 ({\color{cmred}+2.9})
         \\
        
	\bottomrule
	\end{tabular}

\label{tab:com}
\end{table*}

\subsection{Implementation Detail}


\vspace{0.1cm}
\noindent{\bf Souce model pre-training.}
We adopt the DeiT-base network~\cite{touvron2021training} as the backbone of the source pre-trained model, training it in a supervised manner using the source data and corresponding ground truths. 
During this source training phase, the adopted objective is the classic cross-entropy loss with label smoothing, the same as other methods~\cite{liang2020we, yang2021nrc, xu2021cdtrans}.  

\vspace{0.1cm}
\noindent{\bf Variational autoencoder setting.}
The VAE model is an eight-layer convolutional architecture with a latent space dimension of 10.   
We do not employ a pre-trained VAE and utilize a VAE without fine-tuning on any other dataset, ensuring that the learning component ${F}_{\Phi}\left(\frac{\partial \delta_0}{\partial x} \right)$ are unbiased and independent of prior pre-training data.

\noindent\textbf{Parameter setting.} 
For the trade-off parameters in Eq.~\eqref{eqn:loss-all}, we set $\alpha$ to 1.0, while $\beta$ is tuned with \{0.0001, 0.01, 1\} to ensure that the loss values of $L_{\text{KL}}$ and $L_{\text{MSE}}$ remain on the same scale.


\noindent\textbf{Training setting.}
We adopt the batch size of 64, SGD optimizer with a momentum of 0.9 and a learning rate of 1e-5 on all datasets. 
All experiments are conducted with PyTorch on a single GPU of RTX A6000.

\subsection{Comparison Settings}
\paragraph{Evaluation metrics.}
To account for unbalanced datasets, in addition to conventional classification accuracy (termed ACC), we adopt the measure of Quadratic Weighted Kappa (termed QWK)~\cite{qwk} and the average of QWK and ACC (termed AVG). The computation rules of them are provided in \texttt{Supplementary}.  

\vspace{0.1cm}
\noindent{\bf Competitors.}
We compare \modelshortname with nine existing state-of-the-art adaptation methods divided into three groups. 
(1) {\itshape The first group} involves applying the source model directly to the target domain.
(2) {\itshape The second group} includes five {SFDA} methods SHOT~\cite{liang2020we}, 
NRC~\cite{yang2021nrc}, CoWA~\cite{lee2022confidence}, PLUE~\cite{Litrico_2023_CVPR}, and TPDS~\cite{tang2024source}. 
(3) {\itshape The third group} comprises three typical {TTA} methods: SHOT-IM~\cite{liang2020we}, TENT~\cite{wang2020tent}, and SAR~\cite{niutowards}. 

\begin{figure}[h!]
    \centering
    \begin{subfigure}[t]{0.32\linewidth}
        \centering
        \includegraphics[width=\linewidth,height=0.95\linewidth]{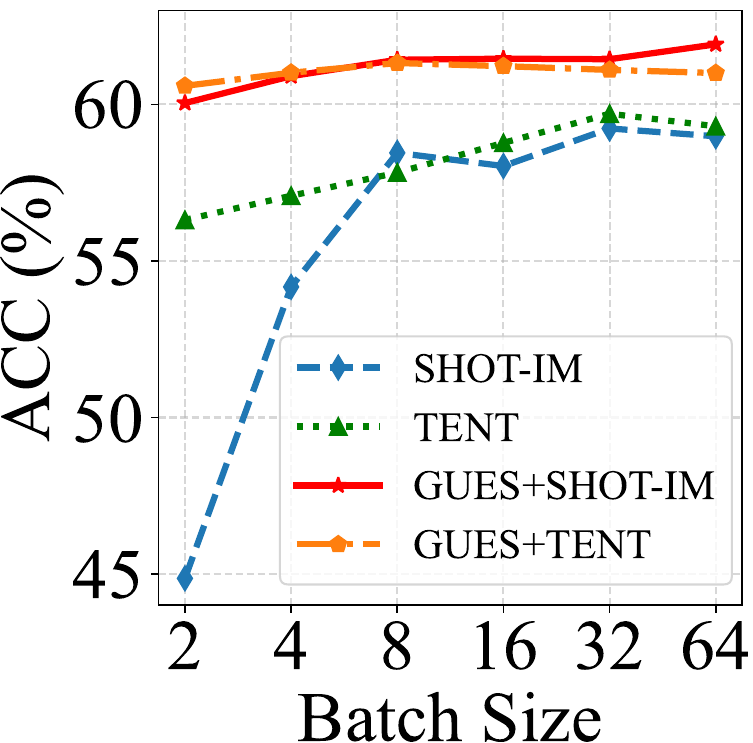}
    \end{subfigure}%
    \begin{subfigure}[t]{0.32\linewidth}
        \centering
        \includegraphics[width=\linewidth,height=0.95\linewidth]{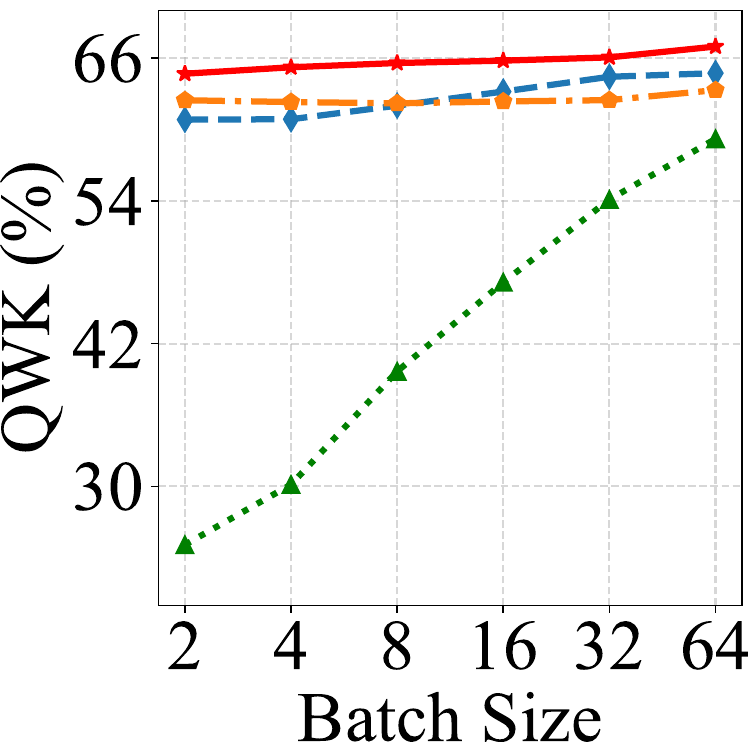}
    \end{subfigure}%
    \begin{subfigure}[t]{0.32\linewidth}
        \centering
        \includegraphics[width=\linewidth,height=0.95\linewidth]{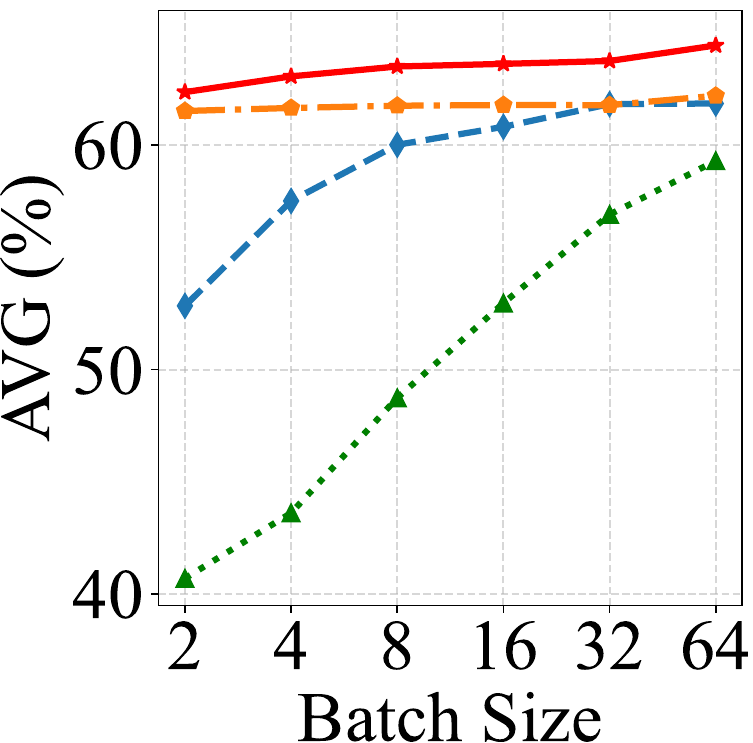}
    \end{subfigure}
    \caption{Comparison results with batch size varying from 2 to 64 over the 12 tasks (The details are provided in \texttt{Supplementary}.).  
    {\bf Left}, {\bf middle}, and {\bf right} report ACC, QWK, and AVG, respectively.
    } 
    \label{fig:batch}
\end{figure}


\vspace{0.1cm}
\noindent{\bf Comparison protocol in {\setting} setting.} 
For a comprehensive comparison, our comparison follows two different fashions as follows.
\begin{itemize}
    \item Case without training: We first generate the unadversarial examples for the target domain by the trained {\modelshortname} model and then provide them to the frozen source model. 
    \item Case with training: We plug {\modelshortname} into other TTA methods (they are also online methods with flowing data) as online image pre-processing. 
\end{itemize}
The two cases evaluate {\modelshortname} from different aspects.  
The first isolates the generalization ability of the unadversarial examples generated by {\modelshortname}, whilst the second highlights {\modelshortname}'s compatibility with other trainable online schemes.

Corresponding to the comparison protocols above, besides the version {\modelshortname} corresponding to the case without training, we also introduce {\modelshortname}+SHOT-IM and {\modelshortname}+TENT which correspond to the case with training.

\begin{figure*}[t]
\setlength{\belowcaptionskip}{0pt}
\setlength{\abovecaptionskip}{0pt}
\begin{center}
    \includegraphics[width=0.9\linewidth]{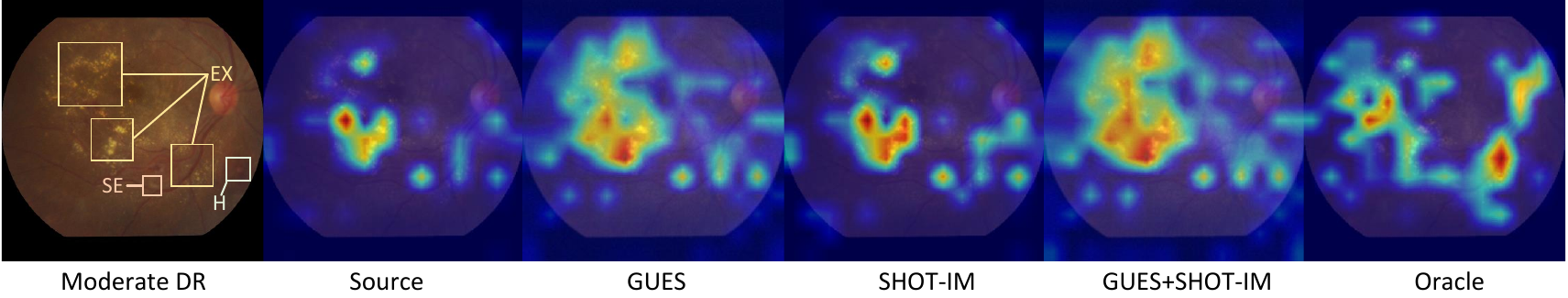}
\end{center}
\caption{Interpretability analysis based on a typical fundus image from ``Moderate DR" class in APTOS. Here, H (hemorrhages), SE (soft exudates), and EX (hard exudates) are essential characteristics to judge the DR grade. 
The gradient CAM-based heatmap of five models visualizes the capture of those lesions. 
All models are trained on task DDR$\to$APTOS, where Oracle is trained using ground truth in APTOS.}

\label{fig:Interpretability}
\end{figure*}

\begin{figure}[t]
    \setlength{\belowcaptionskip}{0pt}
\setlength{\abovecaptionskip}{0pt}
    \centering
    {\includegraphics[width=0.32\linewidth,height=0.32\linewidth]{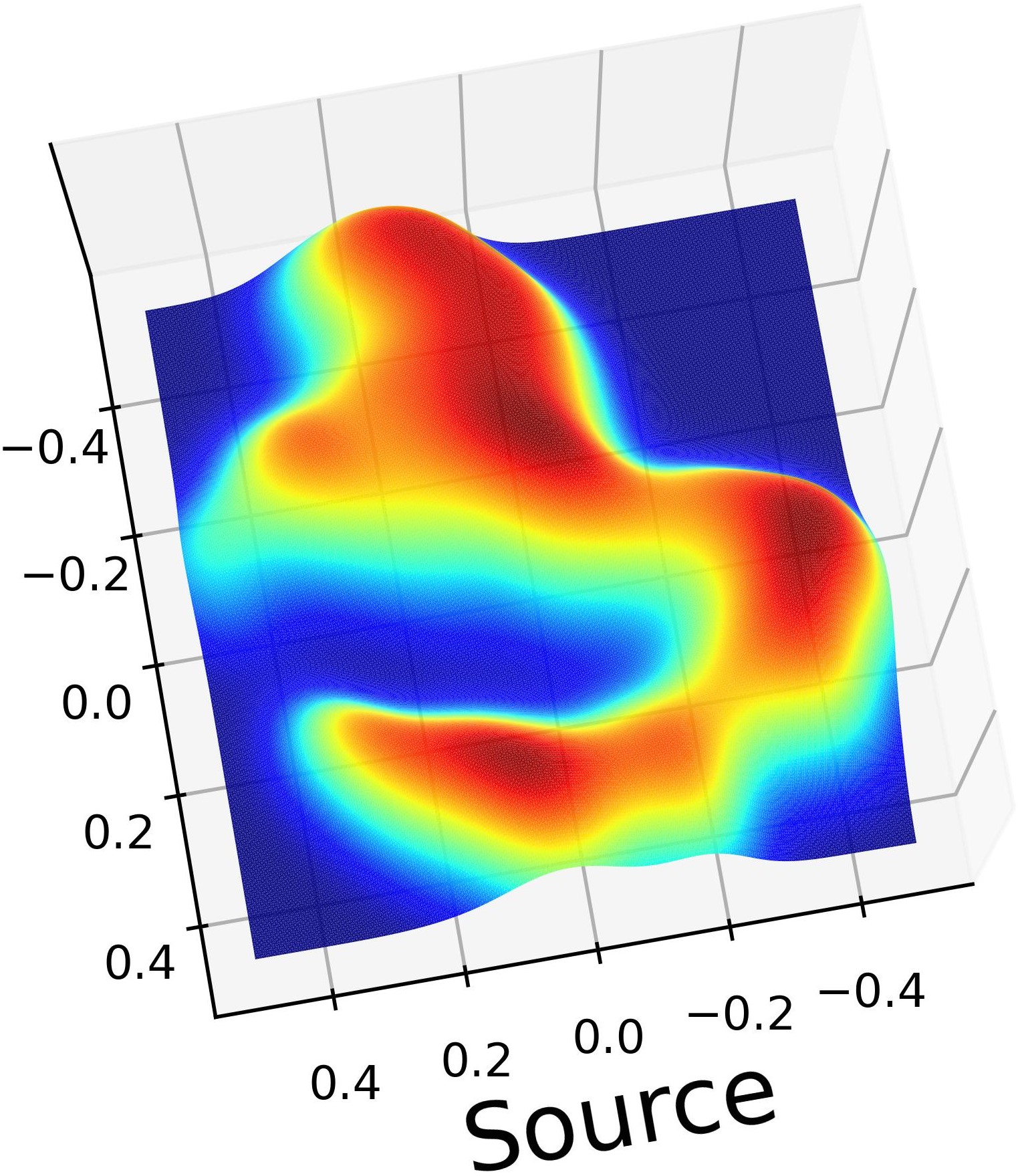}}
    {\includegraphics[width=0.32\linewidth,height=0.32\linewidth]{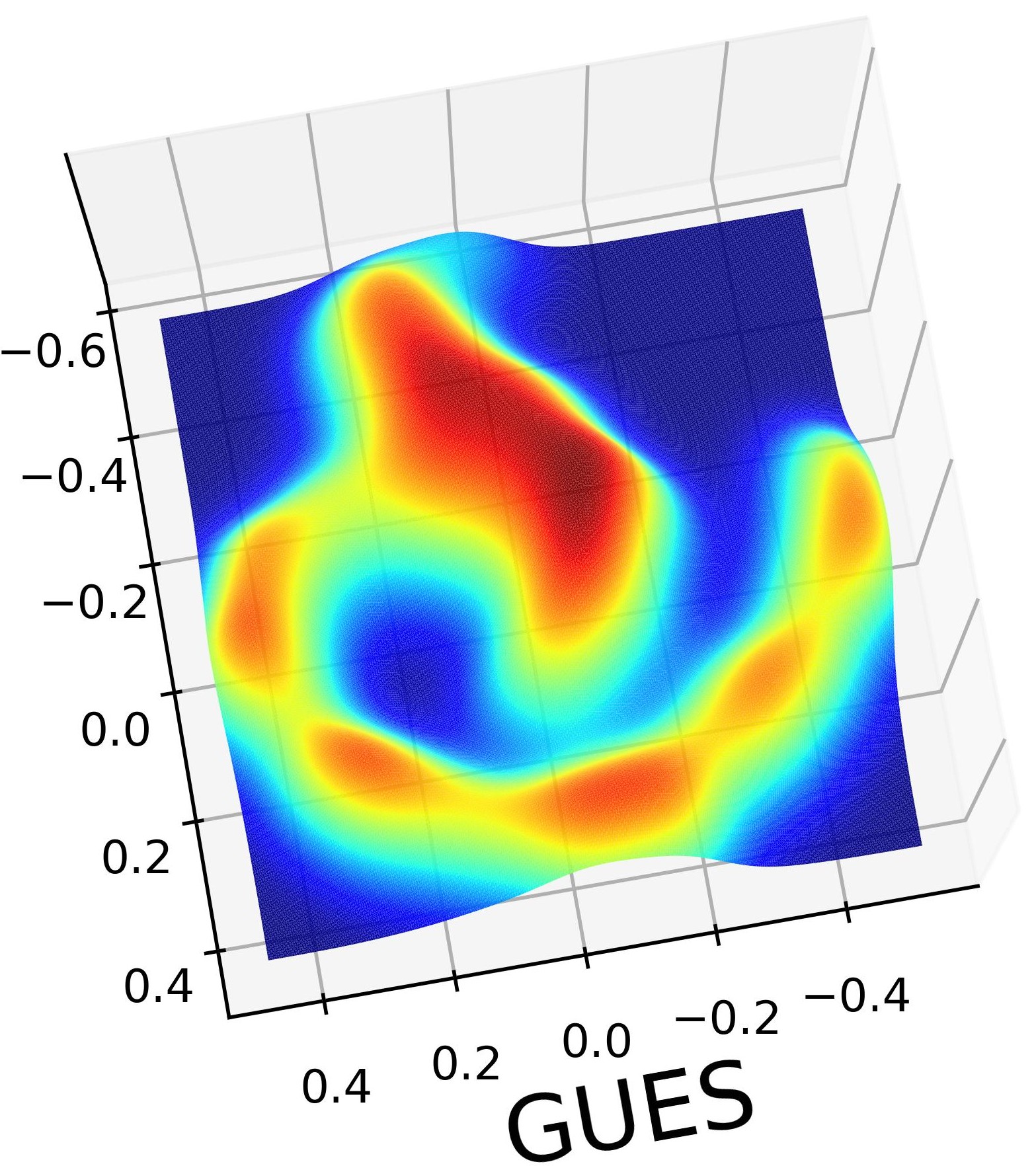}}
    {\includegraphics[width=0.32\linewidth,height=0.32\linewidth]{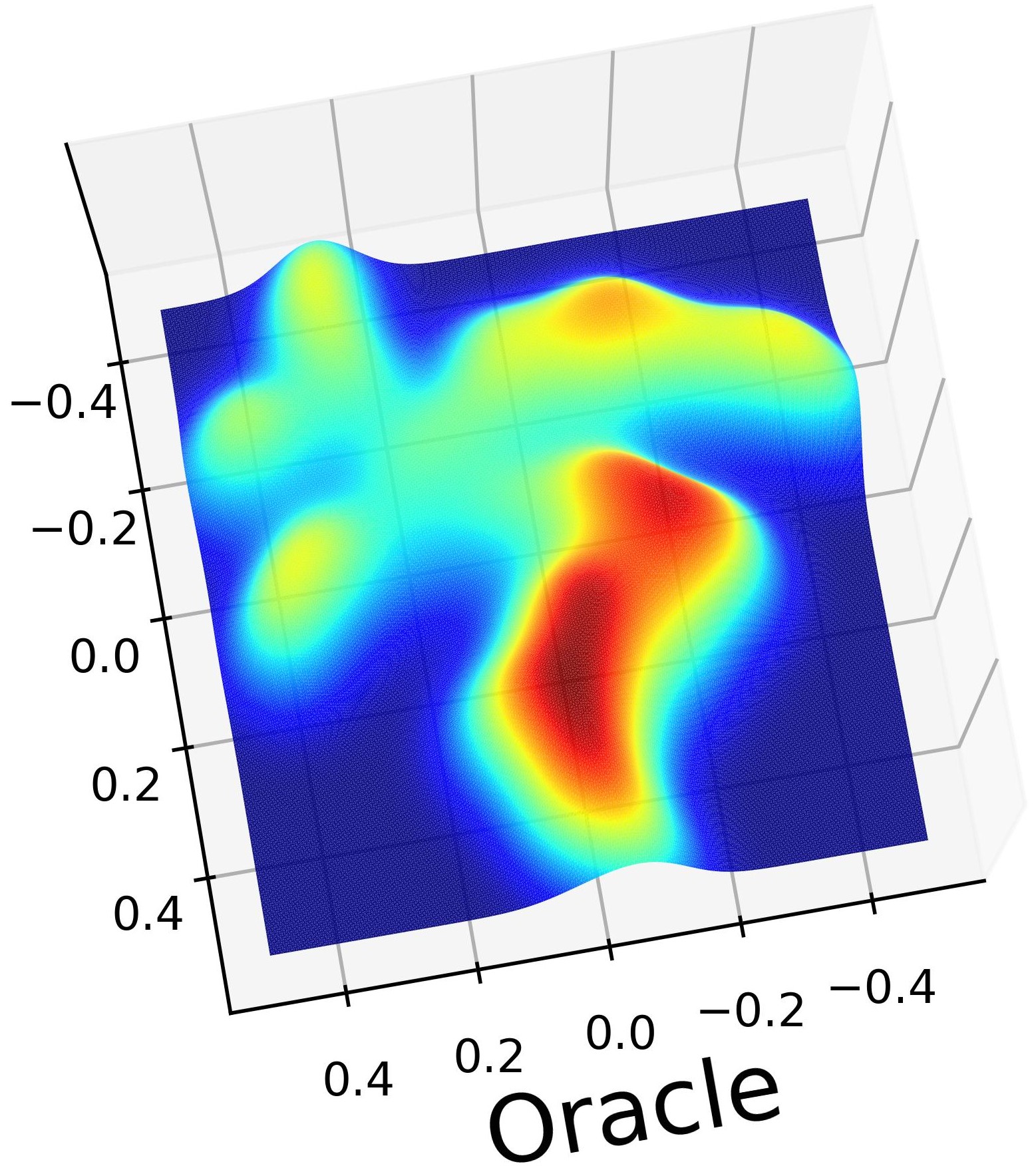}}\\
    \setlength{\abovecaptionskip}{0.3cm}
    \caption{
    Feature distribution comparison of 3D density charts on task DDR$\to$APTOS.  
    Oracle is trained on APTOS by ground truth.}
    \label{fig:visfea}
\end{figure}




\subsection{Comparison Results}
In this part, we present the comparison results following the cases mentioned above. 
Also, considering batch size a crucial factor for TTA methods, the results as the batch size varies is provided.

\vspace{0.1cm}
\noindent{\bf Results without training.}
The comparisons are shown in Tab.~\ref{tab:com}. 
On average, across the 12 tasks and without training of the source model, {\modelshortname} achieves improvements of {\bf 4.5}\% in ACC, {\bf 2.8}\% in QWK, and {\bf 3.7}\% in AVG compared to the source model.  
These results demonstrate that {\modelshortname} modifies the target data distribution effectively, adapting the target domain to align with the source domain.

\vspace{0.1cm}
\noindent{\bf Results with training.}
As shown in Tab.~\ref{tab:com}, {\modelshortname}+SHOT-IM outperforms the previous best SFDA and TTA methods, respectively surpassing TENT in ACC by {\bf 2.7}\%, SHOT in QWK by {\bf 1.9}\%, and SHOT-IM in AVG by {\bf 2.6}\% on average. 
Meanwhile, compared to SHOT-IM, {\modelshortname}+SHOT-IM gains over {\bf 3.0}\% in ACC, {\bf 2.2}\% in QWK, and {\bf 2.6}\% in AVG. 
Similarly, {\modelshortname}+TENT improves over TENT by {\bf 1.7}\% in ACC, {\bf 4.1}\% in QWK, and {\bf 2.9}\% in AVG. These results highlight the effectiveness of combining {\modelshortname} with other methods that require training.

\noindent{\bf Results with varying batch size.}
This part isolates the effect of batch size, which is a crucial factor for TTA methods. 
Fig.~\ref{fig:batch} depicts the performance variation as batch size varying from 2 to 64 over the 12 tasks.
It is observed that TTA methods SHOT-IM and TENT suffer from severe performance drops when the batch size becomes small. 
SHOT-IM exhibits a decrease of approximately {\bf 16}\% in ACC when the batch size is reduced from 64 to 2, whilst TENT shows a substantial decline of around {\bf 34}\% in QWK. 
Oppose to it, the methods with {\modelshortname}, SHOT-IM+{\modelshortname}, and TENT+{\modelshortname}, do not have evident performance decline at the smaller batch size. Moreover, this combination not only mitigates the drop but also shows improvements when the batch size is 64. 
This indicates that {\modelshortname} effectively stabilizes the performance of SHOT-IM and TENT, enhancing their robustness to variations in batch size while boosting their overall effectiveness. 

We attribute {\modelshortname}'s excellent robustness to its ability to predict individual perturbations (see \texttt{Supplementary} for more details on the visualization of perturbations) that focus on single image-specific features rather than global data commonalities. 
For instance, the conventional unadversarial examples approach refines the class-specific perturbation sensitive to batch size. 
Furthermore, both SHOT-IM and TENT are entropy-based methods that require large-scale batch size for accurate entropy estimation.

\subsection{Visualization Analysis}
\paragraph{Interpretability.}
For a better understanding, Fig.~\ref{fig:Interpretability} demonstrates whether {\modelshortname} can help capture pathologically relevant features, such as H (hemorrhages), SE (soft exudates), and EX (hard exudates), determining DR grade. 
First of all, when comparing the source model with GUES, the source model only captures a limited area of the lesion, while {\modelshortname} effectively captures most of the DR-related features. Furthermore, combining {\modelshortname} with SHOT-IM (i.e., {\modelshortname}+SHOT-IM) expands the focus on DR-related features beyond those captured by SHOT-IM alone. Additionally, when comparing the four models to Oracle, only GUES and GUES+SHOT-IM resemble Oracle, suggesting that GUES effectively directs the model's attention to DR-critical features.
\begin{figure}[t]
\setlength{\belowcaptionskip}{0pt}
\setlength{\abovecaptionskip}{0pt}
\begin{center}
    \includegraphics[width=1.0\linewidth]{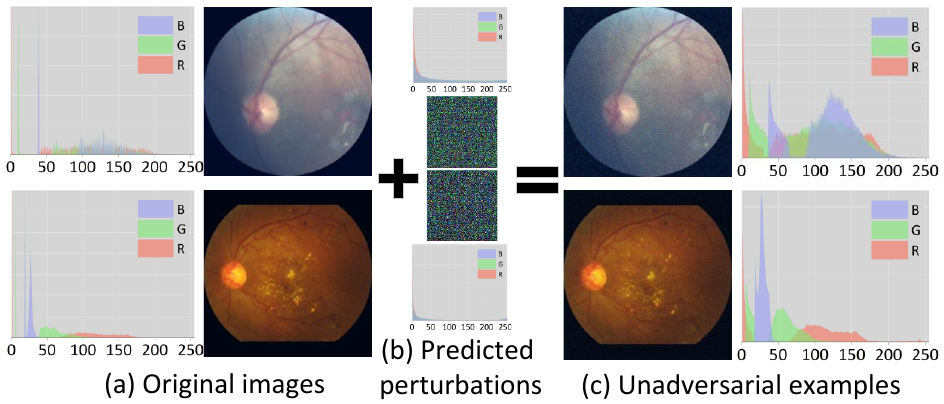}
\end{center}
\caption{Unadversarial examples visualization of two typical target samples from APTOS. The generative perturbations are generated by the {\modelshortname} model trained on task DDR$\to$APTOS.} 
\label{fig:sample}
\end{figure}

\vspace{0.1cm}
\noindent{\bf Feature distribution.}
Taking the task DDR$\to$APTOS as a toy experiment, we visualize the feature distribution extracted from the final convolutional layer of the prediction model using a 3D density chart. 
Considering that APTOS is a class-imbalanced dataset, with the ``No DR" class alone comprising up to {\bf 49.2}\% of the dataset, our analysis focuses on this crucial property.  
As shown in Fig.~\ref{fig:visfea}, the feature distribution of the source model does not reflect this imbalanced characteristic; instead, it displays a more uniform classification. Conversely, the feature distribution of {\modelshortname} exhibits a distinct imbalance, with one expanded high-density region alongside several smaller high-density regions, resembling the distribution pattern seen in Oracle. 

\vspace{0.1cm}
\noindent{\bf Visualization of unadversaisal examples.}
This part visualizes unadversarial examples of two typical target samples from APTOS and corresponding generative perturbations, based on the task DDR$\to$APTOS.
Considering the generative perturbations alter the original images that may not be easily visible to the naked eye, we collect RGB statistics to illustrate these changes quantitatively.
It is observed that in Fig.~\ref{fig:sample}, each channel (R, G, and B) exhibits notable fluctuations, with the RGB statistics of the original images (a) differing significantly from those of the unadversarial examples (c).Additionally, each generative perturbation is unique, meaning that the alterations introduced by these perturbations are individual. These results suggest that the perturbations may help highlight critical DR-related features, refining the model’s focus on diagnostically relevant areas.

\begin{table}[t]
    \centering
    \renewcommand\tabcolsep{5.8pt}
    \renewcommand\arraystretch{1.0}
    \scriptsize
    \label{tab:ab}
    \caption{ACC results of ablation study~(\%).}

        \begin{tabular}{c|cc|cccc|ccc}
        \toprule
        \# &{$L_{\rm{KL}}$} &{$L_{\rm{MSE}}$} &{\bf APTOS} &{\bf DDR} &{\bf DeepDR} &{\bf MD2} &{Avg.} \\
        \midrule
        1&\xmark  &\xmark  &{51.0} &{59.1} &{52.5} &{53.0} &{53.9} \\
        2&\cmark  &\xmark  &54.7 &58.3 &54.9 &53.3 &55.3\\
        3&\xmark  &\cmark  &56.9 &60.1 &52.9 &52.9 &55.7 \\
        4&\cmark  &\cmark  &\textbf{\color{cmblu}60.6} &\textbf{\color{cmblu}61.3} &\textbf{\color{cmblu}56.0} &\textbf{\color{cmblu}55.7} &\textbf{\color{cmblu}58.4}  \\
        \midrule
        5&\multicolumn{2}{l}{\textbf{\modelshortname} w/ ${\rm{AE}}$} \vline &53.8 &60.2 &53.8 &53.2 &55.3\\
        6&\multicolumn{2}{l}{\textbf{\modelshortname} w/ Mixup}     \vline &{56.8} &{60.2} &{53.0} &{52.8} &{55.7} \\
        7&\multicolumn{2}{l}{\textbf{\modelshortname} w/ Self}     \vline &{56.5} &{60.4} &{53.7} &{54.4} &{56.3} \\
        8&\multicolumn{2}{l}{\textbf{\modelshortname} w/ Sal}     \vline &{46.1} &{34.3} &{41.8} &{45.9} &{42.0} \\
        
        \midrule
    \end{tabular} 
    \label{tab:ab_loss}
\end{table}

\subsection{Further Analysis}
\paragraph{Ablation study.}
In this part, we evaluate the effect of objective loss, as well as the components involved in {\modelshortname} including the sampling strategy and saliency map-based supervision.  
To address the first issue, we conduct a progressive experiment. 
The top four rows of Tab.~\ref{tab:ab_loss} list the ablation results where the source model's performance is the baseline. 
Using $L_{\rm{KL}}$ or $L_{\rm{MSE}}$ alone yields an average ACC improvement of approximately {\bf 1.4}\% and {\bf 1.8}\%, respectively, over the baseline. 
As both of them work, the ACC increases {\bf 3.5}\% on average further. 
The results indicate that all objective components positively affect the final performance.  


To evaluate the impact of the sampling component, we propose a variation method of {\modelshortname}, {\modelshortname} w/ ${\rm{AE}}$, where we remove this sampling process by replacing VAE with a conventional Auto-encoder model. 
In addition, two {\modelshortname} variations are used to assess the advantage of saliency map-based supervision. 
Specifically, {\modelshortname} w/ ${\rm{Mixup}}$ replaces the saliency maps with a Mixup of saliency maps and original images, whilst {\modelshortname} w/ ${\rm{Self}}$ replaces the saliency maps with the original images.  
As presented in the Tab.~\ref{tab:ab_loss}, compared with the full version of {\modelshortname} (the fourth row), {\modelshortname} w/ ${\rm{AE}}$, {\modelshortname} w/ ${\rm{Mixup}}$ and {\modelshortname} w/ ${\rm{Self}}$ decrease by {\bf 3.1}\% at least on average. Besides, replacing the original images with saliency maps as inputs ({\modelshortname} w/ ${\rm{Sal}}$) leads to a significant drop of about {\bf 15.4}\%. Above these experiments confirm the effectiveness of our design choices.

\begin{figure}[t]
    \centering
    \begin{subfigure}[t]{0.30\linewidth}
        \centering
        \includegraphics[width=\linewidth,height=1\linewidth]{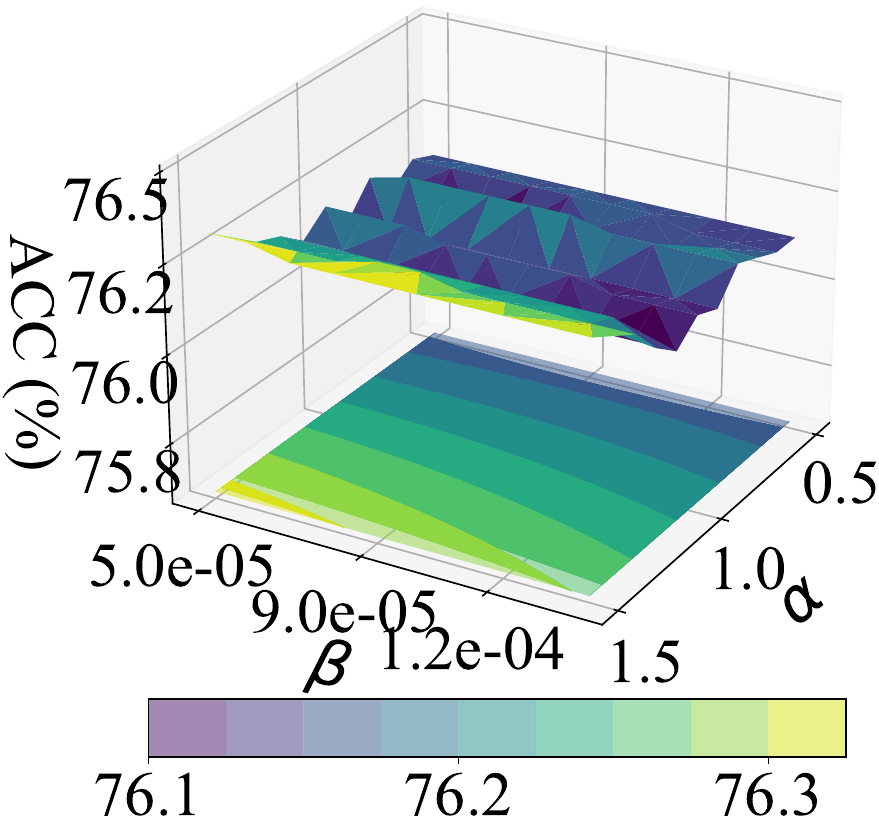}
    \end{subfigure}%
    \hspace{0.1cm}
    \begin{subfigure}[t]{0.30\linewidth}
        \centering
    \includegraphics[width=\linewidth,height=1\linewidth]{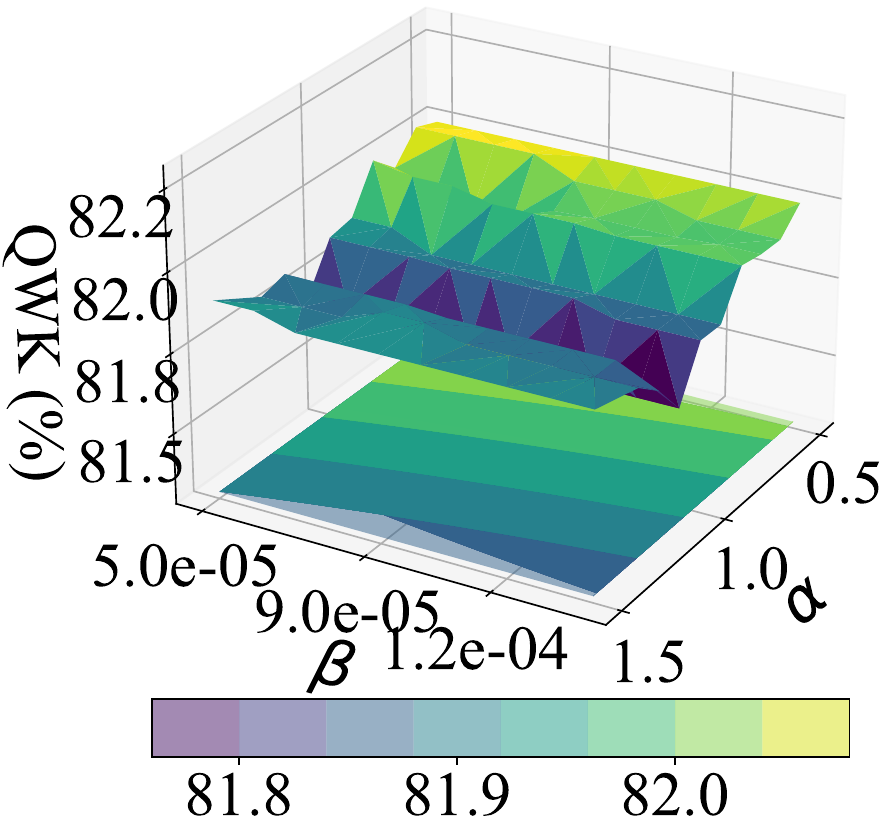}
    \end{subfigure}%
        \hspace{0.1cm}
    \begin{subfigure}[t]{0.30\linewidth}
        \centering
        \includegraphics[width=\linewidth,height=1\linewidth]{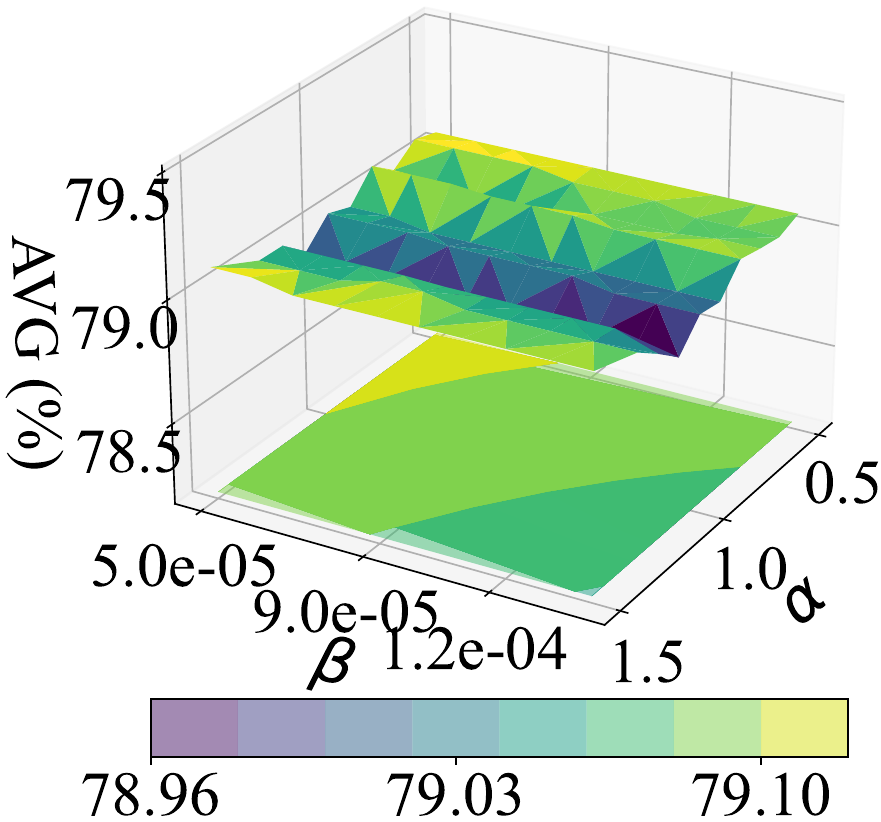}
    \end{subfigure}
    \caption{
Parameter sensitiveness study results over $\alpha \times \beta$ based on task DDR$\to$APTOS. From left to right, there are results on ACC, QWK, and AVG, respectively.} 
    \label{fig:pars-sensi}
\end{figure}

\vspace{0.1cm}
\noindent{\bf Parameter sensitiveness.}
Taking the task DDR$\to$APTOS as a toy experiment, we present the {\modelshortname} performance varying as hyper-parameters $0.5 \le \alpha \le 1.5$ with 0.1 steps, $0.00005 \le \beta \le 0.00014$ with 0.00001 steps. 
As depicted in Fig.~\ref{fig:pars-sensi}, the ACC, QWK, and AVG variation surfaces show fluctuations in the tiny performance zone, with approximately {\bf0.2}\% in ACC, {\bf0.25}\% in QWK, and {\bf0.4}\% in AVG. This observation suggests that {\modelshortname} is insensitive to alterations in $\alpha$ and $\beta$.

\vspace{0.1cm}
\noindent{\bf Limitation.} 
{\modelshortname} uses the saliency map to guide the learning of the generative function. This method is effective for DR grading but encounters challenges in natural image scenarios. 
The natural images contain rich semantics, such as shape, relative structure, and complex background, which are not all relevant to tasks.  
However, the saliency map blindly highlights all those factors, struggling to capture the task-specific ones.  
In a theoretical point-view, the richness of these semantics means significant variations, resulting in a super-relaxed bound constant $U$ (see Theorem~\ref{thm-two}) that undermines the descriptive power of the saliency map for $\frac{\partial \delta_0}{\partial x}$. 
In contrast, fundus images are more monolithic, implying a smaller $U$ that justifies the usage of the saliency map. 
(The further discussion is provided in \texttt{Supplementary})


\section{Conclusion}
In this paper, we propose a clinically motivated setting, {\setting}, where the models are unseen prior to their use, and only target data flows are accessible. 
This setting ensures both model protection and source data privacy in a data flow scenario. 
To adapt to the target domain without access to the models, we introduce a {\modelshortname} approach. 
Instead of conventional iterative optimization, we generate unadversarial examples for flowing target data by directly predicting individual perturbations. 
This approach is grounded in the theoretical results of generative unadversarial learning. 
In practice, we utilize the VAE model to learn the perturbation generation function with a latent input variable. 
Furthermore, we demostrate that saliency maps can serve as an upper bound for this latent variable. 
This relationship inspires us to use the saliency maps as pseudo-perturbation labels for model training.
Extensive experiments conducted on four DR benchmarks confirm that the proposed method can achieve state-of-the-art results, when it pairs with both frozen pre-trained and fine-tuning models.

{
    \small
    \bibliographystyle{ieeenat_fullname}
    \bibliography{main}

\begin{thebibliography}{41}
\providecommand{\natexlab}[1]{#1}
\providecommand{\url}[1]{\texttt{#1}}
\expandafter\ifx\csname urlstyle\endcsname\relax
  \providecommand{\doi}[1]{doi: #1}\else
  \providecommand{\doi}{doi: \begingroup \urlstyle{rm}\Url}\fi

\bibitem[APT(accessed February 20, 2022)]{APTOS2019}
Aptos: Aptos 2019 blindness detection website.
\newblock \url{https://www.kaggle.com/c/aptos2019-blindness-detection}, accessed February 20, 2022.

\bibitem[qwk(accessed July 2022)]{qwk}
Quadratic weighted kappa.
\newblock \url{https://www.Eyepacs.com/aroraaman/ quadratic-kappa-metric-explained-in-5-simple-steps}, accessed July 2022.

\bibitem[AbdelMaksoud et~al.(2020)AbdelMaksoud, Barakat, and Elmogy]{abdelmaksoud2020comprehensive}
Eman AbdelMaksoud, Sherif Barakat, and Mohammed Elmogy.
\newblock A comprehensive diagnosis system for early signs and different diabetic retinopathy grades using fundus retinal images based on pathological changes detection.
\newblock \emph{Computers in Biology and Medicine}, 126:\penalty0 104039, 2020.

\bibitem[Atwany and Yaqub(2022)]{atwany2022drgen}
Mohammad Atwany and Mohammad Yaqub.
\newblock Drgen: domain generalization in diabetic retinopathy classification.
\newblock In \emph{MICCAI}, 2022.

\bibitem[Che et~al.(2023)Che, Cheng, Jin, and Chen]{che2023towards}
Haoxuan Che, Yuhan Cheng, Haibo Jin, and Hao Chen.
\newblock Towards generalizable diabetic retinopathy grading in unseen domains.
\newblock In \emph{MICCAI}, 2023.

\bibitem[Dai et~al.(2021)Dai, Wu, Li, Cai, Wu, Kong, Liu, Wang, Hou, Liu, et~al.]{dai2021deep}
Ling Dai, Liang Wu, Huating Li, Chun Cai, Qiang Wu, Hongyu Kong, Ruhan Liu, Xiangning Wang, Xuhong Hou, Yuexing Liu, et~al.
\newblock A deep learning system for detecting diabetic retinopathy across the disease spectrum.
\newblock \emph{Nature communications}, 12\penalty0 (1):\penalty0 3242, 2021.

\bibitem[Decenci{\`e}re et~al.(2014)Decenci{\`e}re, Zhang, Cazuguel, Lay, Cochener, Trone, Gain, Ord{\'o}{\~n}ez-Varela, Massin, Erginay, et~al.]{decenciere2014feedback}
Etienne Decenci{\`e}re, Xiwei Zhang, Guy Cazuguel, Bruno Lay, B{\'e}atrice Cochener, Caroline Trone, Philippe Gain, John-Richard Ord{\'o}{\~n}ez-Varela, Pascale Massin, Ali Erginay, et~al.
\newblock Feedback on a publicly distributed image database: the messidor database.
\newblock \emph{Image Analysis \& Stereology}, pages 231--234, 2014.

\bibitem[He et~al.(2020)He, Li, Li, Wang, and Fu]{he2020cabnet}
Along He, Tao Li, Ning Li, Kai Wang, and Huazhu Fu.
\newblock Cabnet: Category attention block for imbalanced diabetic retinopathy grading.
\newblock \emph{IEEE Transactions on Medical Imaging}, 40\penalty0 (1):\penalty0 143--153, 2020.

\bibitem[Huang et~al.(2024)Huang, Lyu, Cheng, Tam, and Tang]{huang2024ssit}
Yijin Huang, Junyan Lyu, Pujin Cheng, Roger Tam, and Xiaoying Tang.
\newblock Ssit: Saliency-guided self-supervised image transformer for diabetic retinopathy grading.
\newblock \emph{IEEE Journal of Biomedical and Health Informatics}, 2024.

\bibitem[Kingma(2013)]{kingma2013auto}
Diederik~P Kingma.
\newblock Auto-encoding variational bayes.
\newblock \emph{arXiv:1312.6114}, 2013.

\bibitem[Kouw and Loog(2018)]{kouw2018introduction}
Wouter~M Kouw and Marco Loog.
\newblock An introduction to domain adaptation and transfer learning.
\newblock \emph{arXiv:1812.11806}, 2018.

\bibitem[Lee et~al.(2022)Lee, Jung, Yim, and Yoon]{lee2022confidence}
Jonghyun Lee, Dahuin Jung, Junho Yim, and Sungroh Yoon.
\newblock Confidence score for source-free unsupervised domain adaptation.
\newblock In \emph{ICML}, 2022.

\bibitem[Li et~al.(2019)Li, Gao, Wang, Guo, Liu, and Kang]{li2019diagnostic}
Tao Li, Yingqi Gao, Kai Wang, Song Guo, Hanruo Liu, and Hong Kang.
\newblock Diagnostic assessment of deep learning algorithms for diabetic retinopathy screening.
\newblock \emph{Information Sciences}, 501:\penalty0 511--522, 2019.

\bibitem[Li et~al.(2021)Li, Bo, Hu, Kang, Liu, Wang, and Fu]{li2021applications}
Tao Li, Wang Bo, Chunyu Hu, Hong Kang, Hanruo Liu, Kai Wang, and Huazhu Fu.
\newblock Applications of deep learning in fundus images: A review.
\newblock \emph{Medical Image Analysis}, 69:\penalty0 101971, 2021.

\bibitem[Liang et~al.(2020)Liang, Hu, and Feng]{liang2020we}
Jian Liang, Dapeng Hu, and Jiashi Feng.
\newblock Do we really need to access the source data? source hypothesis transfer for unsupervised domain adaptation.
\newblock In \emph{ICML}, 2020.

\bibitem[Litrico et~al.(2023)Litrico, Del~Bue, and Morerio]{Litrico_2023_CVPR}
Mattia Litrico, Alessio Del~Bue, and Pietro Morerio.
\newblock Guiding pseudo-labels with uncertainty estimation for source-free unsupervised domain adaptation.
\newblock In \emph{CVPR}, 2023.

\bibitem[Liu et~al.(2022)Liu, Wang, Wu, Dai, Fang, Yan, Son, Tang, Li, Gao, et~al.]{liu2022deepdrid}
Ruhan Liu, Xiangning Wang, Qiang Wu, Ling Dai, Xi Fang, Tao Yan, Jaemin Son, Shiqi Tang, Jiang Li, Zijian Gao, et~al.
\newblock Deepdrid: Diabetic retinopathy—grading and image quality estimation challenge.
\newblock \emph{Patterns}, 3\penalty0 (6), 2022.

\bibitem[Liu et~al.(2023)Liu, Kuang, Lin, Wu, and Ji]{liu2023cat}
Xingbin Liu, Huafeng Kuang, Xianming Lin, Yongjian Wu, and Rongrong Ji.
\newblock Cat: Collaborative adversarial training.
\newblock \emph{arXiv:2303.14922}, 2023.

\bibitem[Montabone and Soto(2010)]{montabone2010human}
Sebastian Montabone and Alvaro Soto.
\newblock Human detection using a mobile platform and novel features derived from a visual saliency mechanism.
\newblock \emph{Image and Vision Computing}, 28\penalty0 (3):\penalty0 391--402, 2010.

\bibitem[Nguyen et~al.(2021)Nguyen, Mai, Than, Prange, and Sonntag]{nguyen2021self}
Duy~MH Nguyen, Truong~TN Mai, Ngoc~TT Than, Alexander Prange, and Daniel Sonntag.
\newblock Self-supervised domain adaptation for diabetic retinopathy grading using vessel image reconstruction.
\newblock In \emph{KI}, 2021.

\bibitem[Niu et~al.(2023)Niu, Wu, Zhang, Wen, Chen, Zhao, and Tan]{niutowards}
Shuaicheng Niu, Jiaxiang Wu, Yifan Zhang, Zhiquan Wen, Yaofo Chen, Peilin Zhao, and Mingkui Tan.
\newblock Towards stable test-time adaptation in dynamic wild world.
\newblock In \emph{ICLR}, 2023.

\bibitem[Qiu et~al.(2024)Qiu, Huang, Huang, Yu, and Tang]{qiu2024augpaste}
Jiaming Qiu, Weikai Huang, Yijin Huang, Nanxi Yu, and Xiaoying Tang.
\newblock Augpaste: A one-shot approach for diabetic retinopathy detection.
\newblock \emph{Biomedical Signal Processing and Control}, 96:\penalty0 106489, 2024.

\bibitem[Ran et~al.(2024)Ran, Zhang, Xia, Zhang, Xie, and Zhang]{ran2024source}
Jinye Ran, Guanghua Zhang, Fan Xia, Ximei Zhang, Juan Xie, and Hao Zhang.
\newblock Source-free active domain adaptation for diabetic retinopathy grading based on ultra-wide-field fundus images.
\newblock \emph{Computers in Biology and Medicine}, 174:\penalty0 108418, 2024.

\bibitem[Salman et~al.(2021)Salman, Ilyas, Engstrom, Vemprala, Madry, and Kapoor]{salman2021unadversarial}
Hadi Salman, Andrew Ilyas, Logan Engstrom, Sai Vemprala, Aleksander Madry, and Ashish Kapoor.
\newblock Unadversarial examples: Designing objects for robust vision.
\newblock In \emph{NeurIPS}, 2021.

\bibitem[Sharma et~al.(2023)Sharma, Munz, and Narayan]{sharma2023nsa}
Abhijith Sharma, Phil Munz, and Apurva Narayan.
\newblock Nsa: Naturalistic support artifact to boost network confidence.
\newblock In \emph{IJCNN}, 2023.

\bibitem[Shokri et~al.(2017)Shokri, Stronati, Song, and Shmatikov]{shokri2017membership}
Reza Shokri, Marco Stronati, Congzheng Song, and Vitaly Shmatikov.
\newblock Membership inference attacks against machine learning models.
\newblock In \emph{S\&P}, 2017.

\bibitem[Singer et~al.(1992)Singer, Nathan, Fogel, and Schachat]{singer1992screening}
Daniel~E Singer, David~M Nathan, Howard~A Fogel, and Andrew~P Schachat.
\newblock Screening for diabetic retinopathy.
\newblock \emph{Annals of Internal Medicine}, 116\penalty0 (8):\penalty0 660--671, 1992.

\bibitem[Szegedy(2013)]{szegedy2013intriguing}
C Szegedy.
\newblock Intriguing properties of neural networks.
\newblock \emph{arXiv:1312.6199}, 2013.

\bibitem[Tang et~al.(2024{\natexlab{a}})Tang, Chang, Zhang, Zhu, Ye, and Zhang]{tang2024source}
Song Tang, An Chang, Fabian Zhang, Xiatian Zhu, Mao Ye, and Changshui Zhang.
\newblock Source-free domain adaptation via target prediction distribution searching.
\newblock \emph{International Journal of Computer Vision}, 132\penalty0 (3):\penalty0 654--672, 2024{\natexlab{a}}.

\bibitem[Tang et~al.(2024{\natexlab{b}})Tang, Su, Ye, Zhang, and Zhu]{tang2024unified}
Song Tang, Wenxin Su, Mao Ye, Jianwei Zhang, and Xiatian Zhu.
\newblock Unified source-free domain adaptation.
\newblock \emph{arXiv:2403.07601}, 2024{\natexlab{b}}.

\bibitem[Tomar et~al.(2024)Tomar, Chandel, and Bhatnagar]{tomar2024visual}
Nishtha Tomar, Sushmita Chandel, and Gaurav Bhatnagar.
\newblock A visual attention-based algorithm for brain tumor detection using an on-center saliency map and a superpixel-based framework.
\newblock \emph{Healthcare Analytics}, 5:\penalty0 100323, 2024.

\bibitem[Touvron et~al.(2021)Touvron, Cord, Douze, Massa, Sablayrolles, and J{\'e}gou]{touvron2021training}
Hugo Touvron, Matthieu Cord, Matthijs Douze, Francisco Massa, Alexandre Sablayrolles, and Herv{\'e} J{\'e}gou.
\newblock Training data-efficient image transformers \& distillation through attention.
\newblock In \emph{ICML}, 2021.

\bibitem[Valanarasu et~al.(2024)Valanarasu, Guo, Vibashan, and Patel]{valanarasu2024fly}
Jeya Maria~Jose Valanarasu, Pengfei Guo, VS Vibashan, and Vishal~M Patel.
\newblock On-the-fly test-time adaptation for medical image segmentation.
\newblock In \emph{MIDL}, 2024.

\bibitem[Venkateswara et~al.(2017)Venkateswara, Eusebio, Chakraborty, and Panchanathan]{venkateswara2017deep}
Hemanth Venkateswara, Jose Eusebio, Shayok Chakraborty, and Sethuraman Panchanathan.
\newblock Deep hashing network for unsupervised domain adaptation.
\newblock In \emph{CVPR}, 2017.

\bibitem[Wang et~al.(2020)Wang, Shelhamer, Liu, Olshausen, and Darrell]{wang2020tent}
Dequan Wang, Evan Shelhamer, Shaoteng Liu, Bruno Olshausen, and Trevor Darrell.
\newblock Tent: Fully test-time adaptation by entropy minimization.
\newblock In \emph{ICLR}, 2020.

\bibitem[Wei et~al.(2024)Wei, Huang, Lin, Cheng, Li, and Tang]{wei2024saliency}
Tianyunxi Wei, Yijin Huang, Li Lin, Pujin Cheng, Sirui Li, and Xiaoying Tang.
\newblock Saliency-guided and patch-based mixup for long-tailed skin cancer image classification.
\newblock \emph{arXiv:2406.10801}, 2024.

\bibitem[Wu et~al.(2020)Wu, Shi, Chen, Shi, Chen, Coatrieux, Yang, Luo, and Li]{wu2020coarse}
Zhan Wu, Gonglei Shi, Yang Chen, Fei Shi, Xinjian Chen, Gouenou Coatrieux, Jian Yang, Limin Luo, and Shuo Li.
\newblock Coarse-to-fine classification for diabetic retinopathy grading using convolutional neural network.
\newblock \emph{Artificial Intelligence in Medicine}, 108:\penalty0 101936, 2020.

\bibitem[Xu et~al.(2021)Xu, Chen, Pichao, Wang, Li, and Jin]{xu2021cdtrans}
Tongkun Xu, Weihua Chen, WANG Pichao, Fan Wang, Hao Li, and Rong Jin.
\newblock Cdtrans: Cross-domain transformer for unsupervised domain adaptation.
\newblock In \emph{ICLR}, 2021.

\bibitem[Yang et~al.(2021)Yang, van~de Weijer, Herranz, Jui, et~al.]{yang2021nrc}
Shiqi Yang, Joost van~de Weijer, Luis Herranz, Shangling Jui, et~al.
\newblock Exploiting the intrinsic neighborhood structure for source-free domain adaptation.
\newblock In \emph{NeurIPS}, 2021.

\bibitem[Yin et~al.(2021)Yin, Mallya, Vahdat, Alvarez, Kautz, and Molchanov]{yin2021see}
Hongxu Yin, Arun Mallya, Arash Vahdat, Jose~M Alvarez, Jan Kautz, and Pavlo Molchanov.
\newblock See through gradients: Image batch recovery via gradinversion.
\newblock In \emph{CVPR}, 2021.

\bibitem[Zhang et~al.(2022)Zhang, Lei, and Chen]{zhang2022diabetic}
Chenrui Zhang, Tao Lei, and Ping Chen.
\newblock Diabetic retinopathy grading by a source-free transfer learning approach.
\newblock \emph{Biomedical Signal Processing and Control}, 73:\penalty0 103423, 2022.

\end{thebibliography}
}
\maketitlesupplementary

\section{Reproducibility Statement} 
The code and data will be made available after the publication of this paper.

\section{Proof of Theorem} 
\subsection{A Proof of Theorem 1} 
\noindent\textbf{Recalling traditional unadversarial learning.} 
Unadversarial learning aims to develop an image perturbation that enhances the performance on a specific class, which can be succinctly described as follows:
\begin{equation}
    \label{eqn:unadv}
    \begin{aligned}
        \hat{\delta} = \arg\min \limits_{\delta} {L}({f}_{\theta}({x+\delta}), y), s.t.~||\delta|| \leq \epsilon\\
    \end{aligned}
\end{equation} 
where ${L}\left(\cdot\right)$ denotes objective function, $x$ and $y$ are input image and its label, ${f}_{\theta}$ is a pre-trained model with parameters $\theta$, $\delta$ is a perturbation, $\epsilon$ is a small threshold. Solves this problem in an iterative way formulated as 
\begin{equation}
    \label{eqn:solve-intera}
    \small
    \begin{aligned}
        \delta_{k+1} = \delta_{k} + \alpha\cdot{\rm{sign}}\left(\nabla_{x}{L}({f}_{\theta}({x+\delta_{k}}), y)\right),
        k \in [0, K-1], 
    \end{aligned} 
\end{equation}
where $\alpha$ is a trade-off parameter, $K$ is iteration number, $\delta_0$ is an initial random noise. We re-consider the iterative optimization process above and obtain the theorem below.

\vspace{0.2cm}
\noindent\textbf{Restatement of Theorem~1} 
\textit{Given the unadversarial learning problem defined in Eq.~\eqref{eqn:unadv}, the iterative process featured by Eq.~\eqref{eqn:solve-intera} can be expressed as the following generative form.} 
\begin{equation}
    \label{eqn:theo1}
    {\delta}_{k} = {\delta_0} + V \cdot {F}_{\Phi}\left(\frac{\partial \delta_0}{\partial x} \right), 
\end{equation}
where $\delta_0$ is an initial random noise, $V$ is a bound constant, $F_{\Phi}$ is a generative function.

\vspace{0.2cm}
\noindent{\bf \textit{Proof.}}
First, according to the chain principle, we can convert Eq.~\eqref{eqn:solve-intera} into 
\begin{equation}
    \label{eqn:iteration-chain}
    \begin{aligned}
    \delta_{k+1} &= \delta_{k} + \alpha \cdot \left(\frac{\partial L}{\partial f_{\theta}} \cdot \frac{\partial f_{\theta}}{\partial x} \cdot \left(1 + \frac{\partial \delta_{k}}{\partial x}\right)\right).
    \end{aligned} 
\end{equation}
Since that the learning will converge to the unadversarial examples, $\alpha \cdot \frac{\partial L}{\partial f_{\theta}} \cdot \frac{\partial f_{\theta}}{\partial x}$ is bounded by a certain constant, denoted by $U_k>0$, thereby Eq.~\eqref{eqn:iteration-chain} become 
\begin{equation}
    \label{eqn:iteration-chain-1}
    \begin{aligned}
    \delta_{k+1} \leq \delta_{k} + U_k\left(1 + \frac{\partial \delta_{k}}{\partial x}\right).
    \end{aligned} 
\end{equation}

We make a further substitution on $\delta_{k}$ according to the law presented in Eq.~\eqref{eqn:iteration-chain-1}, leading to 
\begin{equation}
    \label{eqn:iteration-chain-1-1}
    \small
    \begin{aligned}
    \delta_{k+1} &\leq \left[\delta_{k-1} + U_{k-1}\left(1 + \frac{\partial \delta_{k-1}}{\partial x}\right)\right] + U_k\left(1 + \frac{\partial \delta_{k}}{\partial x}\right).
    \end{aligned} 
\end{equation}
By continuing this substitution on $\delta_{k-1}, \cdots, \delta_0$ in order, we have 
\begin{equation}
    \label{eqn:iteration-chain-2}
    \begin{aligned}
    \delta_{k+1} &\leq \delta_{0} + U_{0}\left(1 + \frac{\partial \delta_{0}}{\partial x}\right) + U_{1}\left(1 + \frac{\partial \delta_{1}}{\partial x}\right) \\
    &+\cdots + U_{i}\left(1 + \frac{\partial \delta_{i}}{\partial x}\right) + \cdots + U_{k}\left(1 + \frac{\partial \delta_{k}}{\partial x}\right)\\
    &\leq \delta_{0} + U_m\left[ k + \frac{\partial \delta_{0}}{\partial x} + \frac{\partial \delta_{1}}{\partial x}+ \cdots+\frac{\partial \delta_{k}}{\partial x}\right],
    \end{aligned} 
\end{equation}
where $U_m = max\{U_{0}, U_{1}, \cdots, U_{k}\}$.

To obtain generative form, we explore the relationships between $\{\frac{\partial \delta_{1}}{\partial x}, \frac{\partial \delta_{2}}{\partial x},\cdots, \frac{\partial \delta_{k}}{\partial x}\}$ and $\frac{\partial \delta_{0}}{\partial x}$, respectively. To this end, we first investigate the relationship between $\frac{\partial \delta_{1}}{\partial x}$ and $\frac{\partial \delta_{0}}{\partial x}$, combining Eq.~\eqref{eqn:iteration-chain-1}.
\begin{equation}
    \label{eqn:derive-1}
    \begin{aligned} 
    \frac{\partial \delta_{1}}{\partial x} \leq \frac{\partial \delta_{0}}{\partial x} + U_1\cdot \frac{\partial}{\partial x} \left( \frac{\partial \delta_{0}}{\partial x} \right) = h_1\left(\frac{\partial \delta_{0}}{\partial x}\right)
    \end{aligned} 
\end{equation}
where, $h_1(\cdot)$ stands for an equivalent function. 
For $\frac{\partial \delta_{2}}{\partial x}$, we have the following equation based on Eq.~\eqref{eqn:iteration-chain-1} and Eq.~\eqref{eqn:derive-1}. 
\begin{equation}
    \label{eqn:derive-2}
    \begin{aligned} 
    \frac{\partial \delta_{2}}{\partial x} &\leq \frac{\partial \delta_{1}}{\partial x} + U_2\cdot \frac{\partial}{\partial x} \left( \frac{\partial \delta_{1}}{\partial x} \right)\\ 
    &= h_1\left(\frac{\partial \delta_{0}}{\partial x}\right) + U_2\cdot \frac{\partial}{\partial x} \left( h_1\left(\frac{\partial \delta_{0}}{\partial x}\right) \right)\\ 
    &= h_2\left(\frac{\partial \delta_{0}}{\partial x}\right)
    \end{aligned} 
\end{equation}
In the recursion way presented by Eq.~\eqref{eqn:derive-1} and Eq.~\eqref{eqn:derive-2}, $\{\frac{\partial \delta_{3}}{\partial x},\cdots, \frac{\partial \delta_{k}}{\partial x}\}$ can be expressed as 
\begin{equation}
    \label{eqn:derive-3}
    \begin{aligned} 
    \frac{\partial \delta_{3}}{\partial x} \leq h_3\left(\frac{\partial \delta_{0}}{\partial x}\right),~\cdots,~    \frac{\partial \delta_{k}}{\partial x} \leq h_k\left(\frac{\partial \delta_{0}}{\partial x}\right)
    \end{aligned} 
\end{equation}
Therefore, substituting Eq.~\eqref{eqn:derive-1},~\eqref{eqn:derive-2} and~\eqref{eqn:derive-3} into Eq.~\eqref{eqn:iteration-chain-2}, we have   
\begin{equation}
    \label{eqn:iteration-chain-F}
    \small
    \begin{aligned}
    \delta_{k+1}
    \leq \delta_{0} + U_m\left[ k + \frac{\partial \delta_{0}}{\partial x} + h_1\left(\frac{\partial \delta_{0}}{\partial x}\right)+ \cdots+h_k\left(\frac{\partial \delta_{0}}{\partial x}\right)\right]. 
    \end{aligned} 
\end{equation}

Let $F_{\Phi}\left(\frac{\partial \delta_{0}}{\partial x}\right)=\left[ k + \frac{\partial \delta_{0}}{\partial x} + h_1\left(\frac{\partial \delta_{0}}{\partial x}\right)+ \cdots+h_k\left(\frac{\partial \delta_{0}}{\partial x}\right)\right]$ and $V$ be a value that makes the equality relationship hold. Eq.~\eqref{eqn:iteration-chain-F} becomes the generative form below. 
\begin{equation}
    \label{eqn:theo1-ts}
    {\delta}_{k} = {\delta_0} + V \cdot {F}_{\Phi}\left(\frac{\partial \delta_0}{\partial x} \right). 
\end{equation}

\subsection{A Proof of Theorem 2} 
\noindent\textbf{Recalling the calculation of the fine-grained saliency map.} 
It calculates saliency by measuring central-surround differences within images.
\begin{equation}
    \small
    \label{eqn:sal}
    \begin{split}
    &\mathrm{G}\left(h,w\right)=\sum_\varsigma\max\left\{\mathrm{cen}\left(h,w\right)-\mathrm{sur}\left(h,w,\varsigma\right),0\right\},\\
    &\mathrm{cen}\left(h,w\right) = I\left(h,w\right),\\
    &\mathrm{sur}\left(h,w,\varsigma\right)
    =\frac{\sum\limits^{h' = \varsigma}_{h' = -\varsigma} \sum\limits^{w' = \varsigma}_{w' = -\varsigma}I(h+h',w+w')-I(h,w)}{(2\varsigma + 1)^2 - 1},
    \end{split}
\end{equation}
where $(h, w)$ is the coordinate of one pixel in grey-scale image (transformed by $x_t$) with its corresponding value denoted as $I(w, h)$, and $\varsigma \in \{1, 3, 7\}$ denotes surrounding values.


\vspace{0.2cm}
\noindent\textbf{Restatement of Theorem~2} 
\textit{Given the partial derivatives of the initial random noise $\delta_0$ w.r.t image $x$ is $\frac{\partial \delta_0}{\partial x}$ and $x$'s saliency map is $s=G(x)$ where $G$ is the computation function of saliency map. We have the following relationship:} 
\begin{equation}
    \label{eqn:theo2}
    \frac{\partial \delta_0}{\partial x} \leq U \cdot s, 
\end{equation}
where $U>0$ is a bound constant. 


\vspace{0.2cm} 
\noindent{\bf \textit{Proof.}}
we treat $s$ as a middle variable, thus $\frac{\partial \delta_0}{\partial x}$ can be expressed as the following equation by the chain law. 
\begin{equation}
    \label{eqn:th2-1}
    \begin{aligned}
    \frac{\partial \delta_0}{\partial x} = \frac{\partial \delta_0}{\partial s} \cdot \frac{\partial s}{\partial x} \leq U \cdot\frac{\partial s}{\partial x},
    \end{aligned} 
\end{equation}
where $U>0$ is a bound constant. In Eq.~\eqref{eqn:th2-1}, the inequality holds because both the initial noise and the specific saliency map are bounded, resulting in the relative changes between them also being restricted. 
In addition, according to the definition of derivative, we have 
\begin{equation}
    \label{eqn:th2-2}
    \begin{aligned}
    \frac{\partial s}{\partial x} = \frac{\partial G(x)}{\partial x} \approx  \frac{ G(x+\triangle_x) - G(x)}{\triangle_x},
    \end{aligned} 
\end{equation}
where $\triangle_x$ is a tiny variation.

\begin{figure}[t]
    \begin{center}
        \includegraphics[width=0.45\linewidth]{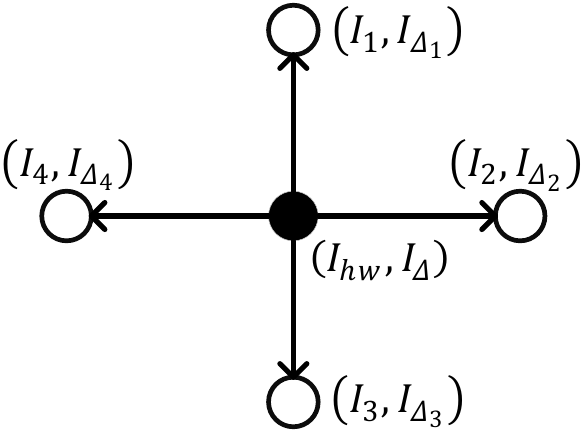}
    \end{center}
    \caption{
    Illustration of $x+\triangle_x$ at coordinate $(h,w)$ as we select the simplest surround case $\varsigma = 1$.
    }
    \label{fig:direct}
\end{figure}

It is known that the saliency map at $(h,w)$ is only related to itself and its surrounding pixels.   
Without loss of generality, we build the proof based on the simplest surround case $\varsigma = 1$ where $\triangle_x$ at $(h,w)$ is presented by Fig.~\ref{fig:direct}. 
According to Eq.~\eqref{eqn:sal}, we have 
\begin{equation}
    \small
    \label{eqn:sur-delta}
    \begin{split}
    &\mathrm{cen}\left(h,w,\triangle_x\right) = \mathrm{cen}\left(h,w\right) + I_{\triangle} = I_{hw} + I_{\triangle}.\\
    &\mathrm{sur}\left(h,w,\varsigma,\triangle_x\right)\\
    &=\frac{\sum_{i=1}^{4}(I_i+I_{\triangle i}) - \left(I_{hw}+I_{\triangle}\right)}{8},\\
    &=\frac{\left(\sum_{i=1}^{4}I_i - I_{hw}\right) + \left(\sum_{i=1}^{4}I_{\triangle i}-I_{\triangle}\right)}{8},\\
    &=\frac{\mathrm{sur}(h,w,\varsigma) + \mathrm{sur}_{\triangle}(\varsigma,\triangle_x)}{8}.
    \end{split}
\end{equation}
Thus, $G(x+\triangle_x)$ at $(h,w)$ can be expressed as 
\begin{equation}
    \small
    \label{eqn:sal-delta}
    \begin{split}
    G_{hw}(x+\triangle_x)&= \sum_\varsigma\max\{\left[\mathrm{cen}\left(h,w\right)-\frac{1}{8}\mathrm{sur}\left(h,w,\varsigma\right)\right]\\
    &-\left[\frac{1}{8}\mathrm{sur}_{\triangle}(\varsigma,\triangle_x)-I_{\triangle}\right],0\} \\
    \end{split}
\end{equation}

Let $A_1 = \mathrm{cen}\left(h,w\right)$ and $A_2 =\mathrm{sur}\left(h,w,\varsigma\right)$, $B_1 = \mathrm{cen}\left(h,w\right)-\frac{1}{8}\mathrm{sur}\left(h,w,\varsigma\right)$, $B_2 = \frac{1}{8}\mathrm{sur}_{\triangle}(\varsigma,\triangle_x)-I_{\triangle}$.
Eq.~\eqref{eqn:th2-2} has two situations as follows. 
\begin{itemize}
\item S-1. When $A_1 > A_2, B_1 > B_2$ or $A_1 < A_2, B_1 > B_2$, 
\begin{equation}
    \small
    \label{eqn:sal-delta-s1}
    \begin{split}
    &\frac{ G(x+\triangle_x) - G(x)}{\triangle_x} \\
    &= \frac{ I_{\triangle} - \frac{1}{8}\mathrm{sur}_{\triangle}(\varsigma,\triangle_x)}{I_{\triangle}}\\
    &=\frac{1}{2}-\sum_{i=1}^{4}\frac{I_{\triangle i}}{I_{\triangle}}
    \end{split}
\end{equation}
\item S-2. When $A_1 > A_2,B_1 < B_2$ or $A_1 < A_2, B_1 < B_2$,
\begin{equation}
    \small
    \label{eqn:sal-delta-s2}
    \begin{split}
    \frac{ G(x+\triangle_x) - G(x)}{\triangle_x}=0
    \end{split}
\end{equation}
\end{itemize}

The results presented above suggest a insight that $\frac{\partial s}{\partial x}$ is 
proportional to the saliency map $s$, namely
\begin{equation}
    \small
    \label{eqn:sal-delta-prop}
    \begin{split}
    \frac{\partial s}{\partial x} \propto s.
    \end{split}
\end{equation}
There are two reasons contributing to this conclusion. 
First, $\frac{\partial s}{\partial x}$' values confine to a binary situation. 
More importantly, as shown in Eq.~\eqref{eqn:sal-delta-s1}, $\frac{\partial s}{\partial x}$ describes the relative change relationship between the current pixel and its surrounding pixels.  
Combing Eq.~\eqref{eqn:th2-1} and Eq.~\eqref{eqn:sal-delta-prop}, we have 
\begin{equation}
    \small
    \label{eqn:sal-delta-final}
    \begin{split}
    \frac{\partial \delta_0}{\partial x} \leq U \cdot \frac{\partial s}{\partial x} \propto U\cdot s.
    \end{split}
\end{equation}

\begin{figure*}[t]
    \begin{center}
        \includegraphics[width=0.9\linewidth]{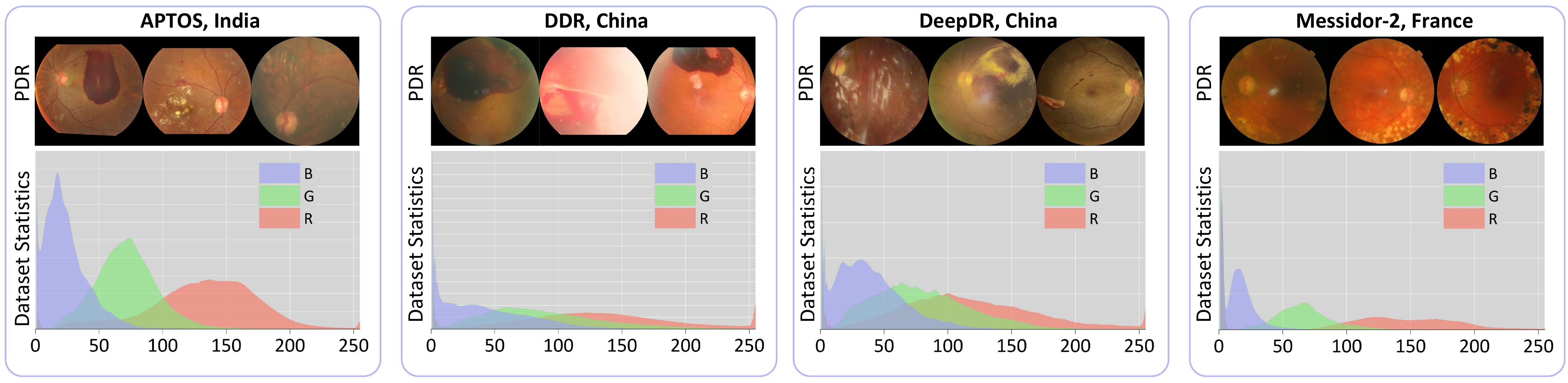}
    \end{center}
    \caption{Visualize the styles and characteristics of each dataset by analyzing the RGB statistics of proliferative diabetic retinopathy (PDR) samples across APTOS, DDR, DeepDR, and Messidor-2.}
    \label{fig:domain}
\end{figure*}

\section{Implementation Details}
\subsection{Datasets Details}
\noindent{\bf{Dataset description.}}
We evaluate the proposed method on four standard DR benchmarks. 
Their details are presented as follows.
\begin{itemize}
    \item \textbf{APTOS.}~\cite{APTOS2019} The dataset originates from Kaggle's APTOS 2019 Blindness Detection Contest, organized by the Asia Pacific Tele-Ophthalmology Society (APTOS). It comprises a total of 5,590 fundus images provided by Aravind Eye Hospital in India. However, only the annotations for the training set (3,662 images) are publicly accessible, and these are used in this study.
    \item \textbf{DDR}~\cite{li2019diagnostic} The DDR dataset comprises 13,673 fundus images collected from 9,598 patients across 23 provinces in China. These images are classified by seven graders based on features such as soft exudates, hard exudates, and hemorrhages.
    \item \textbf{DeepDR}~\cite{liu2022deepdrid} 
    The DeepDR dataset comprises 2,000 fundus images of both left and right eyes from 500 patients in Shanghai, China.
    \item \textbf{Messidor-2}~\cite{decenciere2014feedback} The Messidor-2 dataset includes 1,748 macula-centered eye fundus images. This dataset partially originates from the Messidor program partners, with additional images contributed by Brest University Hospital in France.
\end{itemize}
\begin{table}[t]
    \centering
    \renewcommand\tabcolsep{1.5pt}
    \renewcommand\arraystretch{1.1}
    \scriptsize
    \caption{Label distribution of the four evaluation datasets: \textbf{APTOS}, \textbf{DDR}, \textbf{DeepDR}, and \textbf{Messidor-2}.}
    \begin{tabular}{lllllll}
        \toprule
        Dataset & No DR & Mild DR & Moderate DR & Severe DR& Proliferative DR &Total \\
        \midrule
        \textbf{APTOS}  & 1,805 & 370 & 999 & 193 & 295 &3,662 \\
        \textbf{DDR} &6,265 &630 &4,477 &236&913 &13,673 \\
        \textbf{DeepDR} &914 &222 & 398 & 354  & 112 &2,000\\
        \textbf{Messidor-2} & 1,017 & 270 & 347 & 75 & 35 &1,748\\
        \bottomrule
    \end{tabular}
    \label{tab:data}
\end{table}
\noindent{\bf{The label distribution of datasets.}}
All datasets exhibit imbalanced class distributions, as shown in Table~\ref{tab:data}. Specifically, in APTOS, the ``No DR" class comprises about {\bf 49.2}\% of all samples. In DDR, ``No DR" accounts for approximately {\bf 45.8}\%, while in DeepDR, it makes up around {\bf 45.7}\%. In Messidor-2, the ``No DR" class represents about {\bf 58.2}\% of the total data.
\begin{table*}[t]
\centering
\renewcommand\tabcolsep{3pt}
\renewcommand\arraystretch{0.9}
\scriptsize
\caption{Performance of test time adaptation methods evaluated in ACC, QWK, and AVG across different batch sizes}
\begin{tabular}{lccccccc|ccccccc|ccccccc}  
\toprule
Method & \multicolumn{7}{c|}{ACC} &\multicolumn{7}{c|}{QWK}&\multicolumn{7}{c}{AVG}  \\ \hline
Source & \multicolumn{7}{c|}{53.9} &\multicolumn{7}{c|}{60.1}&\multicolumn{7}{c}{57.0}\\ \hline
&\multicolumn{7}{c|}{Test Time Adaptation Batch Size}  &\multicolumn{7}{c|}{Test Time Adaptation Batch Size}&\multicolumn{7}{c}{Test Time Adaptation Batch Size} \\ \cline{2-22}
          & 2      & 4     & 8     & 16    & 32    & 64    & Avg. & 2      & 4     & 8     & 16    & 32    & 64    & Avg.& 2      & 4     & 8     & 16    & 32    & 64    & Avg.\\ \hline
 SHOT-IM~\cite{liang2020we}  & 44.9	&54.2	&58.5	&58.0	&59.2	&59.0 &55.6 &60.8	&60.9	&62.0	&63.2	&64.4	&64.7	&62.7 &52.8	&57.5	&60.0	&60.8	&61.8	&61.9 &59.1\\ 
 TENT~\cite{wang2020tent}     & 56.3	&57.1	&57.8	&58.8	&59.7	&59.3	&58.2 &25.1	&30.2	&39.7	&47.2	&54.1	&59.2	&42.6 &40.7	&43.6	&48.7	&53.0	&56.9	&59.3 &50.4
  \\ 
\rowcolor{gray! 20} {\textbf{SHOT-IM+GUES}}   &60.0	&60.9	&61.4	&61.5	&61.4	&62.0	&61.2 &64.7	&65.2	&65.6	&65.8	&66.1	&66.9	&65.7 &62.4	&63.1	&63.5	&63.6	&63.7	&64.5 &63.5\\
\rowcolor{gray! 20} {\textbf{TENT+GUES}}   &60.6	&61.0	&61.3	&61.2	&61.1	&61.0	&61.0 &62.5	&62.3	&62.2	&62.4	&62.5	&63.3	&62.5 &61.5	&61.7	&61.8	&61.8	&61.8	&62.2 &61.8\\
\bottomrule
\end{tabular}
    \label{tab:batch}
\end{table*}

\noindent{\bf{The domain shift of datasets.}}
Each dataset is treated as a distinct domain, with significant variations from factors like country of origin, patient demographics, and differences in imaging equipment used for acquisition. Additionally, analysis of the RGB statistics for proliferative DR (PDR) samples across these datasets/domains reveals distinct fluctuations in each channel (R, G, and B), highlighting the unique visual styles and characteristics of each dataset, as shown in Fig.~\ref{fig:domain}.

        


\section{Evaluation metrics.}
The computation rules for accuracy (termed ACC), Quadratic Weighted Kappa (termed QWK), and the average of QWK and ACC (termed AVG) are as follows.
\begin{equation}
    \label{eqn:acc}
    \small
    \begin{aligned}
{ACC}&=\frac{TP+TN}{TP+TN+FP+FN},\\
{QWK}&= 1 - \frac{\sum_{i=1}^{n}\sum_{j=1}^{n} W(i, j) \cdot O(i, j)}{\sum_{i=1}^{n}\sum_{j=1}^{n} W(i, j) \cdot E(i, j)}, {W}_{i,j}=\frac{(i-j)^2}{(C-1)^2}\\
{AVG}&=\frac{1}{2}\left(ACC+QWK\right),
    \end{aligned}
\end{equation}
where $TP$, $TN$, $FP$, and $FN$ represent true positives, true negatives, false positives, and false negatives, respectively. $i$ is a true category, $j$ is a predicted category, $C$ is the number of classes, and $n$ is the total number of samples. $O(i,j)$ is the observed frequency, which represents how many times the true category $i$ was predicted as category $j$, and $E(i,j)$ is the expected frequency, which indicates how many times category $i$ would be predicted as category $j$ under random guessing, $E(i, j) = P(i) \times P(j) \times n.$


\begin{figure*}[t]
\setlength{\belowcaptionskip}{0pt}
\setlength{\abovecaptionskip}{0pt}
\begin{center}
    \includegraphics[width=0.9\linewidth]{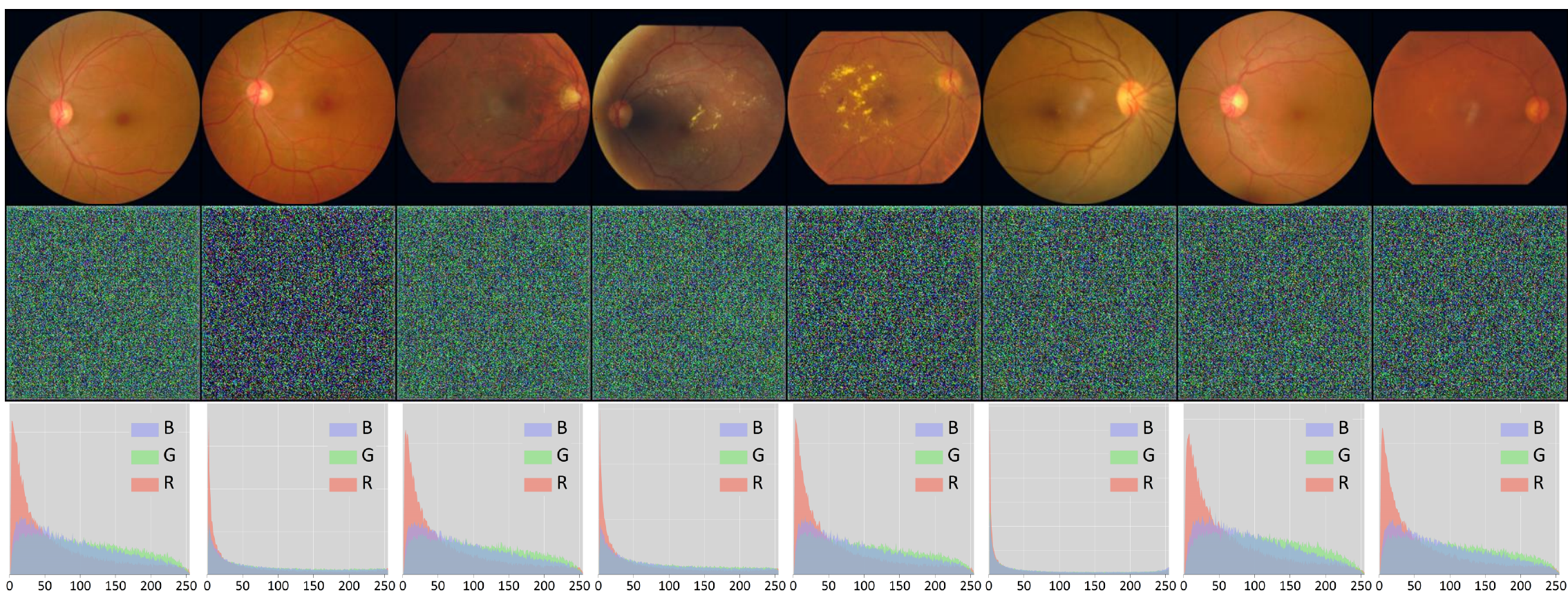}
\end{center}
\caption{Visualization for input images, generative perturbations, and RGB statistic of the corresponding perturbations on transfer task DDR$\to$APTOS. 
}
\label{fig:diff}
\end{figure*}

\begin{figure*}[h!]
\setlength{\belowcaptionskip}{0pt}
\setlength{\abovecaptionskip}{0pt}
\begin{center}
\includegraphics[width=0.88\linewidth]{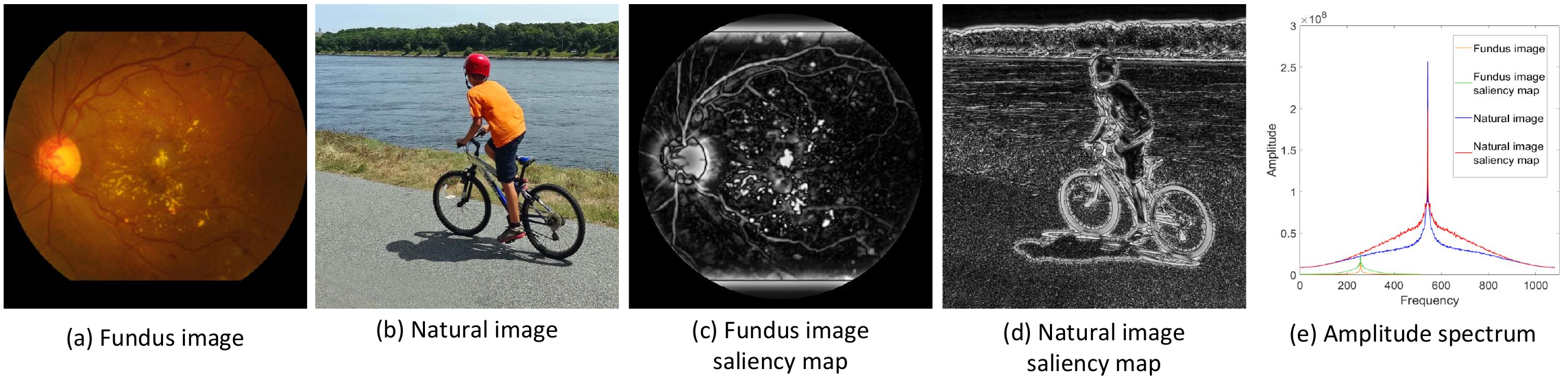}
\end{center}
\caption{Visualization of a fundus image, a natural image, and their corresponding saliency maps. The fundus image is sampled from APTOS, and the natural image is sampled from Office-Home~\cite{venkateswara2017deep}. In (e), the amplitude spectrum of these four images is displayed.
}
\label{fig:natural}
\end{figure*}

\section{\bf Supplementary Experiment Results}
\subsection{Results with Varying Batch Size}
As a supplement to the results with varying batch sizes, Table~\ref{tab:batch} presents the complete performance of three evaluation metrics across all 12 tasks. TTA methods SHOT-IM and TENT show a performance drop when the batch size is small. Specifically, SHOT-IM decreases by approximately {\bf14.1}\% in ACC, {\bf3.9}\% in QWK, and {\bf9.1}\% when comparing batch sizes of 2 and 64. TENT decreases by approximately {\bf3.0}\% in ACC, {\bf34.1}\% in QWK, and {\bf18.6}\% when comparing batch sizes of 2 and 64. However, when these methods are combined with our proposed method, GUES, the decline is not as significant. In SHOT-IM+GUES, the performance shows a decrease of only {\bf2.0}\% in ACC, {\bf2.0}\% in QWK, and {\bf2.1}\% in AVG. In TENT+GUES, the performance shows a decrease of only {\bf0.4}\% in ACC, {\bf0.8}\% in QWK, and {\bf0.7}\% in AVG. These results indicate that our method can prevent declines when the batch size is small, as it predicts individual perturbations that are robust to batch size variations.

\subsection{Visualization for Generative Perturbations.}
As depicted in Fig.~\ref{fig:diff}, it is evident that different input images exhibit distinct perturbations, as observed directly in the second row. To be more specific, the RGB distribution of the perturbations, illustrated in the third row, further highlights their variability. This analysis demonstrates how {\modelshortname} dynamically adjusts the perturbations to account for the unique characteristics of each input image, effectively tailoring them to align with the target domain.

\subsection{Why are Saliency Maps Unsuitable for Natural Images?}
As we early stated, the proposed method cannot tackle the natural image scenarios well. 
This part executes a further discussion for this issue using two typical images illustrated in Fig.~\ref{fig:natural} (a) and (b). 
There are two key observations to note.  
First, the fundus image has a simpler background and structure compared to the natural image, which features richer semantics, including diverse shapes, complex relative structures, and intricate backgrounds. 
This difference is reflected in the amplitude spectrum in Fig.~\ref{fig:natural}(e), where the fundus image displays a significantly lower frequency band. 
Second, the saliency maps effectively highlight variations in both fundus and natural images. This is indicated by the fact that the amplitudes of the saliency maps are much larger than the corresponding amplitudes of the images at similar frequencies.

The effects of this enhancement differ between fundus images and natural images. For simpler fundus images, the noticeable variations are typically related to lesions, making the enhancement useful for highlighting these specific regions (see Fig. \ref{fig:natural} (c)). In contrast, complex natural images exhibit variations that span the entire scene, such as areas of forest, grass, shadows, and a person riding a bike. In this case, the enhancement draws attention to all elements in the image, which can obscure the factors that are relevant to the task at hand. 
Therefore, we believe that refining a proper self-supervised signal for natural images represents a promising research direction for the future.

\end{document}